\documentclass[11pt]{article}

\usepackage{acl}

\usepackage{times}
\usepackage{latexsym}

\usepackage[T1]{fontenc}
\usepackage[utf8]{inputenc}

\usepackage{microtype}

\usepackage{inconsolata}

\usepackage{graphicx}

\usepackage{algorithm}
\usepackage{algorithmic}
\usepackage{amsmath} 

\usepackage{booktabs}
\usepackage{caption}  % Base caption package
\usepackage{subcaption} % For subfigures and subtables
\usepackage{makecell} % Provides \makecell for better control within table cells
\usepackage{array}  
\usepackage{amsmath}
\usepackage{amsthm}
\usepackage{array}
\usepackage{subcaption}
\usepackage{pgf-pie}
\usepackage{adjustbox}
\usepackage{tikz}
\usepackage{tabularx}
\usepackage{xcolor}
\usepackage{colortbl}
\usepackage{graphicx}
\usepackage[absolute,showboxes]{textpos}
\usepackage{multirow}
\usepackage{enumitem}
\setlist{nosep}

\usepackage{latexsym}
\usepackage{placeins}
\usepackage{microtype}
\usepackage{inconsolata}
\usepackage{pgfplots}
\usepackage{pgfplotstable}
\usepackage{pifont}

\definecolor{darkgreen}{RGB}{0,100,0}
\newtheoremstyle{compactexample} % Define a new style for the example environment
    {3pt} % Space above
    {3pt} % Space below
    {} % Body font
    {} % Indent amount
    {\bfseries} % Theorem head font
    {.} % Punctuation after theorem head
    {.5em} % Space after theorem head
    {} % Theorem head spec

\theoremstyle{compactexample}

\usepackage{newfloat}
\usepackage{listings}

\title{Which Words Matter Most in Zero-Shot Prompts?}

\author{%
  Nikta Gohari Sadr$^1$\textnormal{,} 
  Sangmitra Madhusudan$^1$\textnormal{,} \\ 
  \textbf{Hassan Sajjad}$^2$
  \textnormal{and}  
  \textbf{Ali Emami}$^3$\\
  $^1$Brock University, St. Catharines, Canada \\
  $^2$Dalhousie University, Halifax, Canada \\
  $^3$Emory University, Atlanta, USA \\
}

\begin{document}
\maketitle

\begin{abstract}
While zero-shot instructional prompts like ``Let's think step-by-step'' have revolutionized Large Language Model performance, a fundamental question remains unanswered: which specific words drive their remarkable effectiveness? We introduce the ZIP score (Zero-shot Importance of Perturbation), the first systematic method to quantify individual word importance in instructional prompts through controlled perturbations including synonym replacement, co-hyponym substitution, and strategic removal. Our analysis across four flagship models, seven widely-adopted prompts, and multiple task domains reveals four key findings: (1) Task-specific word hierarchies exist where mathematical problems prioritize ``step-by-step'' while reasoning tasks favor ``think''; (2) Proprietary models show superior alignment with human intuitions compared to open-source alternatives; (3) Nouns dominate importance rankings, consistently representing the majority of significant words; and (4) Word importance inversely correlates with model performance, indicating prompts have greatest impact where models struggle most. Beyond revealing these patterns, we establish the first ground-truth benchmark for prompt interpretability through 20 validation prompts with predetermined key words, where ZIP achieves 90\% accuracy versus LIME's 60\%. Our findings advance \textit{prompt science}, the study of how language shapes model behavior, providing both practical insights for prompt engineering and theoretical understanding of word-level effects in LLMs.\footnote{The complete codebase and documentation of language model interactions will be publicly available upon publication.}

\end{abstract}

\section{Introduction}

\begin{figure*}[!ht]
  \centering
  \includegraphics[width=0.77\textwidth]{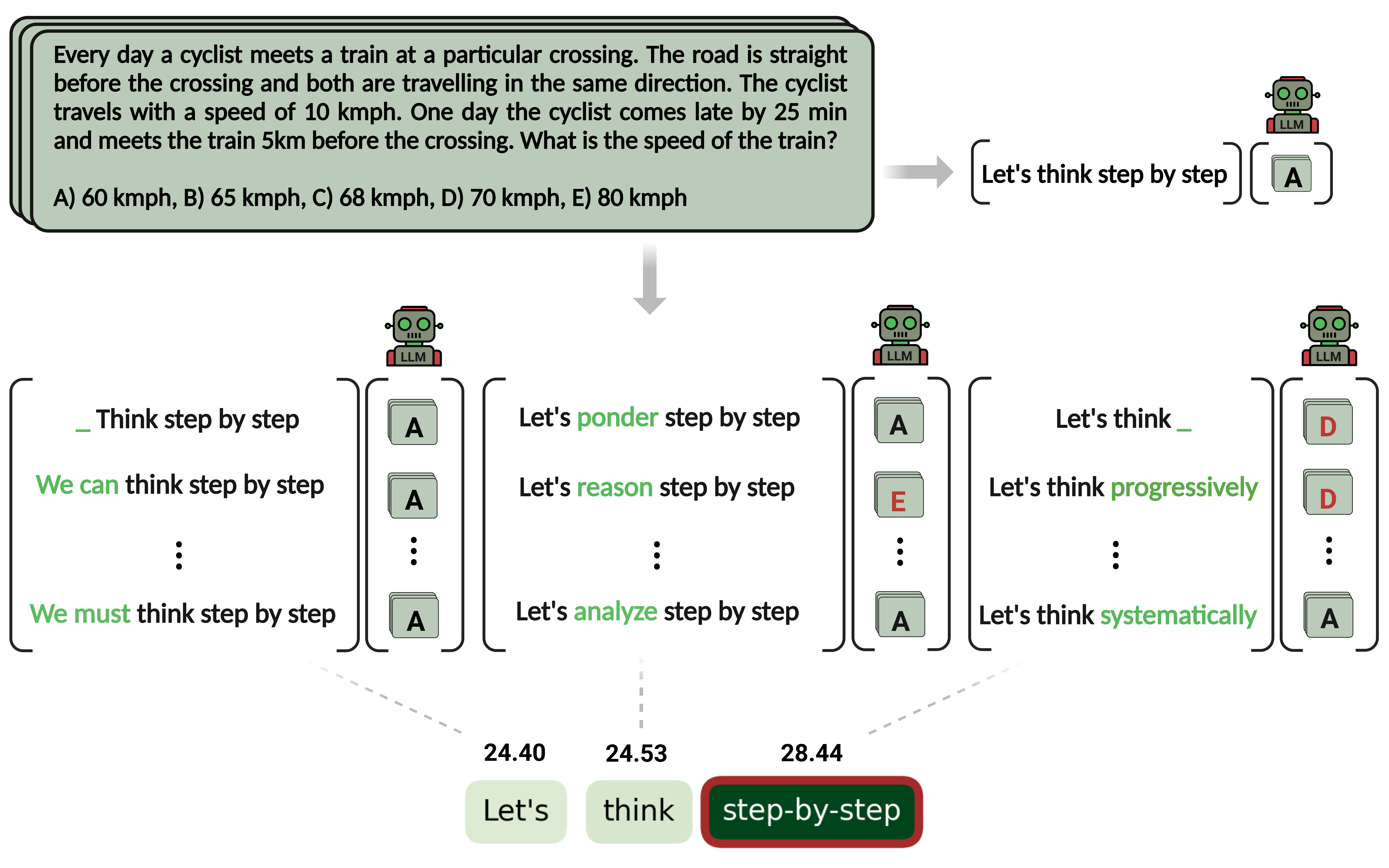}

  \caption{ZIP score generation process for the Chain-of-Thought prompt using GPT-4o mini on a AQUA-RAT dataset instance. Perturbed prompts are compared to the original, generating word-level ZIP scores. The red box highlights ``step-by-step'' as significantly important based on statistical analysis.}
  \label{fig:Process}

\end{figure*}

Chain-of-Thought prompting transformed how we use Large Language Models. Adding ``Let's think step-by-step'' to a prompt, GPT-3 suddenly solved complex math problems it previously failed on \cite{wei2022chain}. This discovery sparked a wave of instruction prompts: Self-Consistency \cite{wang2023selfconsistency}, Plan-and-Solve \cite{wang2023planandsolvepromptingimprovingzeroshot}, and others, each improving performance through carefully crafted phrases. These prompts didn't just boost task performance; they became fundamental to training Large Reasoning Models like OpenAI's o3 \cite{openai2025o3mini}.

Consider the Plan-and-Solve prompt: ``Let's first understand the problem and devise a plan to solve it. Then, solve it step-by-step'' \cite{wang2023planandsolvepromptingimprovingzeroshot}. This manually crafted extension of Chain-of-Thought consistently outperforms the original prompt across benchmarks. The intuition \textit{seems} clear: explicit planning should reduce errors. But which words actually drive this improvement? Is it the instruction to ``plan''? The directive to ``understand''? Or does the familiar ``step-by-step'' do most of the work? Despite the empirical success of these prompts, the selection of which words to include remains largely \textit{ad hoc}, we lack systematic understanding of how individual word choices affect model performance.

While it is well-known that models are sensitive to prompt structure \cite{sclar2024quantifyinglanguagemodelssensitivity}, the importance of individual word choices in a prompt remains poorly understood. Individual word choices can have unexpected and counterintuitive effects on model performance. For instance, replacing ``solve'' with its synonym ``work out'' seems harmless, but can produce vastly different outputs. %(see Appendix \ref{sec:extended_ap}). 

%While it is well-known that models are sensitive to prompt structure, our experiments reveal surprising patterns at the individual word level. Removing ``step-by-step'' from a prompt should hurt performance, yet we find cases where it actually improves accuracy. Replacing ``solve'' with its synonym ``work out'' seems harmless, but can produce vastly different outputs (see Appendix \ref{sec:extended_ap}). These counterintuitive results suggest that individual word choices can have dramatic and unexpected effects on model performance.

Following established work on perturbation-based interpretability \cite{choudhary2022interpretation, SALEEM2022165, feng-etal-2018-pathologies}, we define \textbf{word importance} as \textit{the degree to which a word's presence, absence, or modification impacts model performance on a given task}. While gradient-based \cite{wallace-etal-2019-allennlp, yin2022interpreting, ferrando-etal-2023-explaining, wu2023analyzing} and attention-based methods \cite{modarressi2022globenc, tenney2020language, ferrando-etal-2022-measuring} also aim to identify important words, they infer importance from model internals rather than directly measuring performance impact.

Beyond this, methods that analyze internal representations face practical limitations. They require access to model internals, making them inapplicable to closed-source models like GPT-4. They also face reliability issues: gradients can be manipulated \cite{wang2020gradient} and attention weights often fail to correlate with actual performance impact \cite{jain-wallace-2019-attention, ethayarajh-jurafsky-2021-attention}. Recent work on mechanistic interpretability (i.e., circuit analysis, sparse autoencoders, and activation patching) \cite{lindsey2025biology} provides detailed internal analysis but requires substantial computational resources and whitebox access. For understanding how word choices actually affect model outputs, we need lightweight, model-agnostic methods that directly measure performance changes.

We introduce the ZIP score (Zero-shot Importance of Perturbation score), a metric that quantifies word importance through systematic perturbations including synonym replacement, co-hyponym substitution, and word removal. Our approach works with any model, requiring only the ability to query the model. Additionally, we develop a set of validation prompts with known ground-truth word importance, providing a benchmark for evaluating interpretability methods. Figure~\ref{fig:Process} shows our method identifying ``step-by-step'' as the critical component in Chain-of-Thought prompting for a mathematical reasoning task.

Testing across four flagship models, seven prompts, and multiple tasks reveals surprising patterns. Word importance varies dramatically by task type: ``step-by-step'' dominates in mathematical problems while ``think'' matters more for common-sense reasoning. Proprietary models like GPT-4o mini show closer alignment with human intuitions than open-source alternatives. Across all models, nouns consistently emerge as most important (47.4\%--65.9\% of significant words). Most strikingly, ZIP scores inversely correlate with model performance $(|\text{r}| > 0.9)$: prompts have their greatest impact on tasks where models struggle.

These findings advance \textit{prompt science} \cite{shahpromptscience2025}, the systematic study of how prompts elicit model behavior, by quantifying word level importance in zero-shot instructional prompts. By revealing which words actually matter and how word choices affect model outputs, we provide both practical guidance for prompt design and deeper insights into prompt-model interactions.

\section{Related Work}

Feature importance interpretation in language models is typically approached through three main methods: gradient-based, attention-based, and perturbation-based \cite{choudhary2022interpretation}.

\textbf{Gradient-Based Methods} identify influential features by calculating the gradient of the output logits with respect to input elements \cite{wallace-etal-2019-allennlp}. Recent advances include gradient-based post hoc contrastive explanations for model predictions \cite{yin2022interpreting,ferrando-etal-2023-explaining} and analyses of how CoT prompting affects saliency scores \cite{wu2023analyzing}. However, these methods face significant limitations: they require access to model internals, making them inapplicable to closed-source models. Furthermore, research has shown that model gradients can be easily manipulated, raising concerns about the reliability of gradient-based analyses \cite{wang2020gradient}.

\textbf{Attention-Based Methods} interpret feature importance by analyzing the weighted sum of intermediate representations in neural networks \cite{modarressi2022globenc, tenney2020language, ferrando-etal-2022-measuring}. While intuitive, these approaches have several drawbacks. For one, the attention weights do not always correspond to the most important features for model predictions \cite{jain-wallace-2019-attention} and may not correlate well with other feature importance measures \cite{ethayarajh-jurafsky-2021-attention}. Moreover, attention weights often contain redundant and mixed information, leading to unreliable explanations \cite{bastings-filippova-2020-elephant}. Like gradient-based methods, these approaches also require access to model internals, limiting their applicability to open-source models.

\textbf{Perturbation-Based Methods} can analyze any LLM regardless of architecture accessibility by measuring how outputs change when inputs are modified. As \cite{choudhary2022interpretation} describes, ``a word (token) or a collection of words (tokens) are modified or removed from the input samples, and a resulting change is measured.'' Notable approaches include LIME \cite{ribeiro2016should}, which creates local linear approximations of model behavior, and other dataset instance perturbations \cite{zafar2019dlime}. Recent research has extended these methods to few-shot demonstrations \cite{liu2023towards} and system prompts \cite{hackmann2024word, yin2023did}.

Despite these advancements, there remains a significant gap in the interpretation of zero-shot instructional prompts. Current research often uses basic token masking, which may oversimplify prompt semantics. It can also result in incoherent perturbations, potentially failing to interpret the true impact of the word on model predictions \cite{yin2023did,feng-etal-2018-pathologies}. Our research addresses this by introducing a new metric that employs a variety of meaningful perturbations to reveal the importance of each word in zero-shot instructional prompts.

\textbf{Prompt Science}: Recent work distinguishes between prompt engineering (optimizing prompts for specific tasks) and prompt science (using prompts to discover regularities in model behavior) \cite{holtzman2025prompting,shahpromptscience2025}. Studies have documented prompt brittleness, the sensitivity of model outputs to minor prompt variations \cite{arora2022ask, min-etal-2022-rethinking, sclar2023quantifying}. For instance, \citet{sclar2023quantifying} found that formatting changes alone can swing accuracy by over 70 percentage points. Rather than viewing this sensitivity as a flaw, we leverage it as a tool for understanding word importance. As \citet{holtzman2025prompting} argued, prompt sensitivity reflects models' attempts to ``infer substantial information from limited context,'' making it a feature for scientific investigation rather than a bug to be fixed. Our ZIP score provides a systematic framework for what they call ``behavioral studies that use varied prompts in structured ways to confirm hypotheses.''

\section{The ZIP score}

The ZIP score quantifies the importance of individual words in a prompt by measuring how much the model's performance changes when that word is perturbed or removed. 
%To support broader use, we developed a public interface to visualize word importance in any zero-shot prompt on a given input task. Following AAAI's policy on supplementary material, we provide screenshots of this tool in Appendix \ref{sec:web} rather than a live link. The interface will be made publicly available upon publication along with our complete codebase. %This method allows us to systematically identify which words are most crucial for the model's performance on a given task.

\subsection{Formalization of the ZIP score}

We formalize the ZIP score through the following steps, using a running example:

\text{Example Task ($T$):} Determine whether a number is prime or composite.

 \text{Original Prompt ($P$):} ``Take a deep breath and work on this problem step-by-step'' \cite{luo2024taking}.

\noindent\textbf{Step 1: Prompt Representation}

Let \(P\) represent the original prompt, where \(P = w_1, w_2, \dots, w_I\) and \(w_i\) denotes the \(i\)-th word. Throughout this paper, we use ``word'' to refer to any token resulting from our space-based tokenization method. This may include contractions, compound words, or multi-word expressions.
\text{Example:}
%\[P = \left[ \text{``Take'', ``a'', ``deep'', ``breath'', ``and'',} \right.\]
%\vspace{-5.5mm}
%\[\left. \text{``work'', ``on'', ``this'', ``problem'', ``step-by-step''} \right]\]
\[
P = [\text{``Take''}, \text{``a''}, \text{``deep''},  \ldots, \text{``step-by-step''}]
\]

\begin{figure}[!t]
  \centering
  \includegraphics[width=0.4\textwidth]{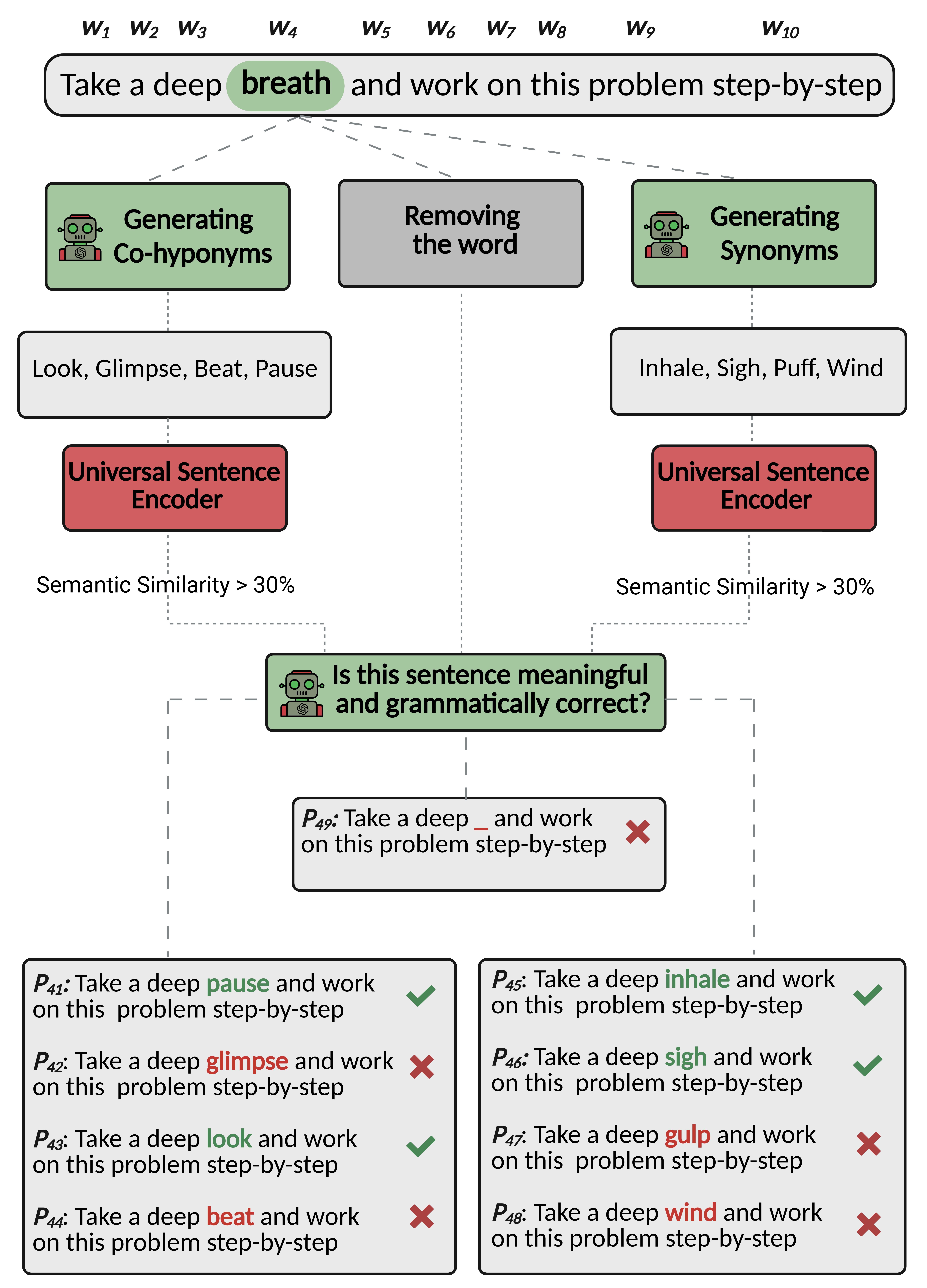}

  \caption{Perturbation generation and filtering pipeline}
  \label{fig:Creating_Perturbations}
\end{figure}

\noindent\textbf{Step 2: Perturbation Generation}

For each word \(w_i\) in \(P\), we generate a set of perturbations \(\{P_{i1}, P_{i2}, \dots, P_{iJ}\}\), through context-aware synonym replacement and co-hyponym replacement using GPT-4. Synonyms provide similar meanings, co-hyponyms introduce related concepts, and removal tests word necessity. Each \(P_{ij}\) represents a modified version of \(P\) where \(w_i\) has been replaced or omitted. %\footnote{GPT-4 consistently outperformed traditional tools like NLTK \cite{bird2009natural} in generating contextually appropriate alternatives in our preliminary tests.}, and word removal. 
%A co-hyponym is a word that shares the same category or group with another word, but both are specific examples of a broader term (hypernym). 

\text{Example:} For \(w_4 = \text{``breath''}\):
\begin{itemize}
    \item \(P_{41} = \text{``Take a deep pause and work on ...''}\)
    \item \(P_{42} = \text{``Take a deep glimpse and work on ...''}\)
    \item \(P_{43} = \text{``Take a deep look and work on  ...''}\)
\end{itemize}

Aligned with existing literature on perturbation-based methods \cite{choudhary2022interpretation, IVANOVS2021228, SALEEM2022165}, we posit that a word’s importance is indicated by the model’s performance shifts in response to any meaningful alteration of that word, whether through subtle meaningful shifts or its complete removal (see \S\ref{sec:proofofthepudding} for analyses and demonstrations).

%We use GPT-4 \cite{achiam2023gpt} for generating context-aware synonyms and co-hyponyms, solely for single-word perturbation generation.\footnote{GPT-4 consistently outperformed traditional tools like NLTK \cite{bird2009natural} in generating contextually appropriate alternatives in our preliminary tests.}

To create a final set of perturbations that provides insight into the impact of each word on the model output, as depicted in Figure \ref{fig:Creating_Perturbations}, we filter the perturbed sentences as follows:

\begin{enumerate}[itemsep=0pt, leftmargin=*]
\item The Universal Sentence Encoder \cite{cer2018universal} to measure semantic similarity between $P$ and $P_{ij}$, keeping those with $>$30\% similarity. This threshold was chosen to balance maintaining context while allowing meaningful alterations (see Table \ref{emperical_test} in \ref{sec:pert_ap}).

\item GPT-4 evaluation to assess grammatical correctness and semantic viability of each $P_{ij}$.%\footnote{In preliminary experiments, we manually verified that GPT-4 performed this task with near-perfect precision and recall in assessing meaningful perturbations.}
\end{enumerate} 
%This process, depicted in Figure \ref{fig:Creating_Perturbations}, aims to produce a final perturbation set that offers meaningful insights into each word's impact on model outputs.

Our filtering process excludes \(P_{42}\) for insufficient semantic similarity and grammatical issues, leaving \(P_{41}\) and \(P_{43}\) as valid perturbations.

We maintained cost-effectiveness by designing concise prompts and compact output formats to minimize token usage, and by using the Universal Sentence Encoder to filter out unrelated perturbations before making API calls. Across the seven instructional prompts, this filtering process resulted in an average of 9.04 perturbed sentences per word. Detailed prompts and examples for perturbation creation and evaluation are provided in Appendix Sections \ref{sec:template_ap} and \ref{sec:pert_ap}.%\footnote{For the CoT prompt, the entire perturbation generation and grammatical verification process uses fewer than 2150 input/output tokens, effectively balancing cost and quality.}.

\noindent\textbf{Step 3: Model Prediction}

We input the original prompt \(P\) and each of its validated perturbed variants \(P_{ij}\) into the language model \(M\) to address an instance \((t)\) of task \(T\).

\text{Example ($t$):} Is 29 prime or composite?
\begin{align*}
\text{pred}_t^M(P) &= \text{``Prime''} \quad \text{(Correct)} \\
\text{pred}_t^M(P_{41}) &= \text{``Prime''} \quad \text{(Correct)} \\
\text{pred}_t^M(P_{43}) &= \text{``Composite''} \quad \text{(Incorrect)}
\end{align*}

\noindent\textbf{Step 4: Disagreement Calculation}

We define a disagreement function \(d\) that measures the difference between model predictions for each perturbed prompt (\(P_{ij}\)) and the original prompt (\(P\)). This function varies by task type \(T\). In our study, for classification tasks:
\[
d_t^M(P, P_{ij}) = 
\begin{cases} 
1 & \text{if } \text{pred}_t^M(P) \neq \text{pred}_t^M(P_{ij}) \\
0 & \text{otherwise}
\end{cases}
\]

For translation tasks:
\[
d_t^M(P, P_{ij}) = |\text{BLEU}_t^M(P) - \text{BLEU}_t^M(P_{ij})|
\]

Here $\text{BLEU}_t^M(P)$ and $\text{BLEU}_t^M(P_{ij})$ are the BLEU scores \cite{papineni2002bleu} for translations generated from the original and perturbed prompts, respectively.

Since the above example \(
T\) is a classification task, we calculate disagreement scores as follows:
% \textbf{Example:}
\begin{align*}
d_t^M(P, P_{41}) &= 0 \quad \text{(No disagreement)} \\
d_t^M(P, P_{43}) &= 1 \quad \text{(Disagreement)}
\end{align*}

%For an instance \(t\) of translation tasks:

%Here $\text{BLEU}_t^M(P)$ and $\text{BLEU}_t^M(P_{ij})$ are the BLEU scores \cite{papineni2002bleu} for translations generated from the original and perturbed prompts, respectively.

%\textbf{Translation Example:} For the task of translating ``The cat is on the mat'' to French, assume:

%$\text{BLEU}_t^M(P) = 0.85$, 

%$\text{BLEU}_t^M(P_{41}) = 0.83$,

%$\text{BLEU}_t^M(P_{43}) = 0.72$

%\noindent $d_t^M(P, P_{41}) %= |0.85 - 0.83| = 0.02$, 
%$d_t^M(P, P_{43}) = |0.85 - 0.72| = 0.13$

\noindent\vspace{2mm}\textbf{Step 5: ZIP score Calculation}

Finally, for each word $w_i$, we compute 
%the ZIP score 
$\text{ZIP}_T^M(w_i)$ by averaging disagreement scores across all \(J\) perturbations of $w_i$ and all dataset instances:
\[
\text{ZIP}_T^M(w_i) = 100 \cdot \frac{1}{N} \sum_{t=1}^{N} \left(\frac{1}{J} \sum_{j=1}^{J} d_t^M(P, P_{ij})\right),
\]

where $N$ is the number of dataset instances, $P$ is the original prompt, and $P_{ij}$ is the $j$-th perturbation of word $w_i$ in the $t$-th dataset instance.%\footnote{Due to the linearity of expectation, the order of averaging across datasets and perturbations can be interchanged without affecting the final ZIP score.}

\text{Example:} For $w_4 = \text{``breath''}$ in a single instance ($N=1$):

\noindent \[ \text{ZIP}_{T}^M (\text{``breath''}) = 100 \cdot \frac{1}{2}(0 + 1) = 50
\]

%\noindent $\text{ZIP}_{Transl.}^M (\text{``breath''}) = 100 \cdot \frac{1}{2}(0.02 + 0.13) = 7.5$

The ZIP score ($\in [0,100]$) represents the average percentage change in model performance observed across all tested dataset instances when perturbing a given word.  This metric is adaptable to various performance measures, making it versatile for different types of tasks. Interpreting the ZIP score is straightforward: a higher score indicates greater word importance, as perturbing it leads to larger changes in model output. Conversely, a lower score suggests the word has less impact on performance for the given task.

\begin{table*}[h]
\centering
\footnotesize % smaller font
\setlength{\tabcolsep}{1pt} % reduce column padding
\renewcommand{\arraystretch}{0.95} % adjust row spacing
\begin{tabularx}{\textwidth}{@{}p{1.1cm}Xc@{}}
\toprule
\textbf{Code} & \textbf{Prompt} & \textbf{Mutual Datasets} \\ 
\midrule
0-CoT & Let's think step-by-step. \cite{10.5555/3600270.3601883} &  BIG-bench, GSM8K, AQUA-RAT \\ 
0-CoTB & Take a deep breath and work on this problem step-by-step. \cite{yang2024largelanguagemodelsoptimizers} &  AQUA-RAT, GSM8K \\ 
0-CoTR & Let’s work this out in a step-by-step way to be sure we have the right answer. \cite{zhou2023largelanguagemodelshumanlevel} & GSM8K, BIG-bench \\ 
0-IRR & Feel free to ignore irrelevant information in the problem description. \cite{shi2023largelanguagemodelseasily} &  GSM8k \\ 
0-PS & Let’s first understand the problem and devise a plan to solve it. Then, solve it step-by-step. (Plan \& Solve \cite{wang2023planandsolvepromptingimprovingzeroshot}) & GSM8K, AQUA-RAT \\ 
0-DSP & Provide the translation step-by-step, then complete the sentence.  \cite{peng2023makingchatgptmachinetranslation} &  WMT19 (German, Chinese) \\ 
0-DTG & Detect the error type first, then refine the translation. (Deliberate then Generate \cite{li2023deliberategenerateenhancedprompting}) & WMT (German, Chinese) \\ 
\bottomrule
\end{tabularx}
\caption{Zero-shot instructional prompts used in our experiments, and mutual datasets they were tested on.}
\label{tab:Prompt_datasets}
\end{table*}

\subsection{Identifying Significantly Important Words}

To determine which words are significantly important in prompts, we use a statistical approach that distinguishes between the inherent variability in model outputs and the effects of our perturbations:
\begin{enumerate}[itemsep=0pt, leftmargin=*]
\item For prompt word, we generate a set of perturbed prompts (the same set used in the ZIP score calculation).

\item We repeatedly input both the original prompt and each of its perturbed versions to the model, generating multiple outputs for each, up to the maximum number of word perturbations in the prompt. 
%The number of times we re-prompt with the original prompt matches the number of perturbations for that word, ensuring a balanced comparison.

\item We compare the ZIP scores from the original \textit{re}-prompts with those from each perturbed prompt using the Wilcoxon rank-sum test.
\end{enumerate}

\noindent The Wilcoxon rank-sum test tests whether two independent samples (i.e., the outputs from the original re-prompts and a perturbed prompt) come from the same distribution. We deem words with a p-value $<$ 0.05 as `significantly important'. %We use a temperature of 0.5 for all prompts to balance determinism and variability in model responses.%\footnote{Setting temperature to 0 didn't eliminate output variations, a known issue with models like GPT-4.} %This approach allows us to distinguish between changes in model output caused by perturbations and those resulting from the model's inherent variability. %By doing so, we can confidently identify words whose perturbations consistently and significantly affect model performance.

\section{Experimental Setup}
\label{sec:exp_setup}

\subsection{Validation Prompts}
\label{sec:validation_prompts}

Despite the success of zero-shot prompts, establishing a ground truth for word-level importance within them is challenging because we may not know which words \textit{should} matter most. To address this, we created a set of 20 simple zero-shot instructional validation prompts, each designed so that a single predetermined `key word' determines the correct output. For example, consider the prompt \textit{``Say the word green''}. Here, ``green'' is the key word: if the model does not produce ``green'' exactly, it fails. By knowing in advance which word is crucial for a correct output, we can objectively assess the effectiveness of interpretability methods.

We constructed 20 validation prompts (Table~\ref{tab:Control_Prompt_4omini}) with varied syntax and structure, such as \textit{``Print the digits \textbf{123}''}, \textit{``Type the letter \textbf{X}''}, and \textit{``Print \textbf{carrot} with no additional text''}. Each prompt contains a key word that directly determines the correct output, any alteration to this word should change the model's response. For example, in ``Print the digits 123'', only perturbations to ``123'' should affect whether the model outputs ``123'' correctly.

\subsection{Prompts}

We evaluated seven instructional prompts across various datasets, as detailed in Table \ref{tab:Prompt_datasets}. These prompts were selected based on their proven effectiveness and generalization capabilities in recent literature. We used a simple space-based tokenization method, where tokens are defined as sequences of characters separated by whitespace. This approach preserves contractions and compound words (e.g., ``Let's'' as one token), which are often semantically important in instructional prompts. Dataset-specific prompts are included in the Appendix \ref{sec:template_ap} (Table \ref{tab:task_prompt}).

\subsection{Datasets}
We selected a diverse range of datasets known for effectively testing zero-shot prompts, as highlighted in bold in Table \ref{tab:Prompt_datasets}. For classification tasks, datasets included GSM8K \cite{cobbe2021trainingverifierssolvemath}, AQUA-RAT \cite{ling2017programinductionrationalegeneration}, and the Big Bench dataset \cite{srivastava2023imitationgamequantifyingextrapolating}. Translation tasks used the WMT19 dataset, concentrating on Chinese and German languages. We evaluated on 150 instances per dataset, which we determined was sufficient to provide statistically significant results while managing computational costs.

\subsection{Models}
Four models were used in our experiments: GPT-4o mini, GPT-3.5-turbo, Llama-2-70B-chat, and Mixtral-8x7B-Instruct-v0.1 \cite{openai2023gpt4, touvron2023Llama, jiang2024mixtral}. Each model was configured with a temperature of 0.5 to balance between deterministic outputs and allowing for some variability. For token usage estimates per prompt, see Appendix \ref{sec:computation_ap}.

\subsection{Baselines}
\label{sec:baselines}

We compare our ZIP method with LIME \cite{ribeiro2016should}, a widely-used perturbation-based explanation method. For validation prompts, we configured LIME to treat the task as binary classification: Label 1 represents the target word (e.g., ``green'' for ``Say the word green''), and Label 2 represents any other output. LIME automatically creates perturbations through random word removal and fits a linear surrogate model to quantify word importance, using 50 perturbations per instance. %Unlike ZIP's diverse perturbation approach (synonyms, co-hyponyms, and removal), LIME relies solely on random word removal to estimate importance through linear modeling. This implementation allows direct comparison with ZIP scores while maintaining LIME's standard approach to interpretability.

\subsection{Human Evaluation}
We conducted a study with 15 diverse participants proficient in English to compare human and model interpretations of the instructional language. For each prompt, participants selected up to three words that they deemed most important to solve a certain task. The evaluation form is shown in Appendix \ref{sec:human_eval_ap} Fig. \ref{fig:human_eval}. We used the Jaccard Index to measure inter-annotator agreement across participants, yielding an average of 0.4714, indicating moderate agreement. Jaccard Index was chosen over Fleiss' Kappa to accommodate multiple word selections per participant.
Table~\ref{tab:human_QA} in the Appendix illustrates the selected words across participants and prompts. While there is moderate agreement overall, the table highlights clear variation in which words individuals judged as most important. This variation points at the subjectivity of human intuition and the need for model-based evaluations.

\section{Results}

\subsection{Validation Results}

Table \ref{tab:Control_Prompt_4omini} compares how ZIP and LIME performed in identifying key words on our 20 validation prompts on GPT-4o mini. ZIP correctly identified the predetermined key word in 90\% of prompts, significantly outperforming LIME's 60\% accuracy. LIME's errors often involved misidentifying contextual words (e.g., ``Print,'' ``Output,'' ``Say") as most important, while ZIP maintained focus on the target words that directly determine the output.

ZIP's strong performance was consistent across all four models (e.g., with 100\% accuracy on GPT-3.5-turbo), validating ZIP's effectiveness in identifying significant words in instructional prompts. Detailed results are provided in Appendix \ref{sec:Validation_prompt_app}.

\subsection{Cross-Model Analysis}

Table  
   \ref{tab:CoT_Heatmap_BB} presents ZIP score heatmaps for the CoT prompt ``Let's Think Step-by-Step'' across models on Big Bench. Our analysis reveals:
\begin{itemize}[itemsep=0pt, leftmargin=*]
      \item \textbf{``Step-by-Step'' vs ``Think'':} While both words are important, \textbf{step-by-step} shows, on average, a higher ZIP score across models. This suggests that the explicit instruction for a structured approach (``step-by-step'') may be more universally interpreted and utilized by models than the more general cognitive instruction (``think'').
\item \textbf{Proprietary vs Open-source models:} Proprietary models (GPT-4o mini, GPT-3.5-turbo) identify fewer words as significantly important (1 word) compared to open-source models (Mixtral, Llama 2, with 2-3 words). This suggests that proprietary models are relatively robust, showing less dependence on single words in prompts, while open-source models exhibit broader sensitivity to individual word choices. %This could reflect differences in training approaches or model architectures in proprietary and open-source models.
\end{itemize}

  \noindent  Results on all datasets are provided in Appendix \ref{sec:cot_heatmap_ap} Table \ref{tab:CoT_Heatmaps_Comparison}.
  
%\begin{table}[t]
%\centering
%\resizebox{\columnwidth}{!}{
%\begin{tabular}{@{}lc@{}} % 'l' for left-aligned text, 'c' for centered images
%\toprule
%GPT-4o mini  & \adjustbox{valign=m, margin=0.7ex 0.7ex 0.7ex 0.7ex}{\includegraphics[width=0.33\textwidth]{Images/cot_heatmap_gpt4_bigbench.png}}\\
%GPT 3.5 Turbo & \adjustbox{valign=m, margin=0.7ex 0.7ex 0.7ex 0.7ex}{\includegraphics[width=0.33\textwidth]{Images/cot_heatmap_gpt35_bigbench.png}}\\
%Mixtral       & \adjustbox{valign=m, margin=0.7ex 0.7ex 0.7ex 0.7ex}{\includegraphics[width=0.33\textwidth]{Images/cot_heatmap_mixtral_bigbench.png}}\\
%Llama 2       & \adjustbox{valign=m, margin=0.7ex 0.7ex 0.7ex 0.7ex}{\includegraphics[width=0.33\textwidth]{Images/cot_heatmap_Llama2_bigbench.png}}\\
%\bottomrule
%\end{tabular}
%}

%\caption{ZIP score heatmaps for the CoT prompt across four models on the Big Bench dataset. Red boxes indicate significantly important words.}
%\label{tab:CoT_Heatmap_BB}
%\end{table}

\begin{table}[t]
\centering
\resizebox{\columnwidth}{!}{
\begin{tabular}{@{}lc@{}} % 'l' for left-aligned text, 'c' for centered images
\toprule
GPT-4o mini  & \adjustbox{valign=m, margin=0.7ex 0.7ex 0.7ex 0.7ex}{\includegraphics[width=0.33\textwidth]{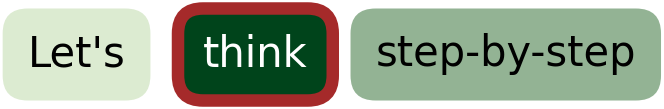}}\\
GPT 3.5 Turbo & \adjustbox{valign=m, margin=0.7ex 0.7ex 0.7ex 0.7ex}{\includegraphics[width=0.33\textwidth]{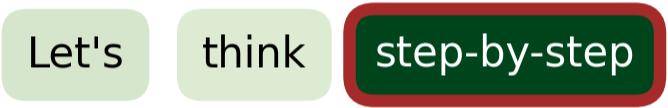}}\\
Mixtral       & \adjustbox{valign=m, margin=0.7ex 0.7ex 0.7ex 0.7ex}{\includegraphics[width=0.33\textwidth]{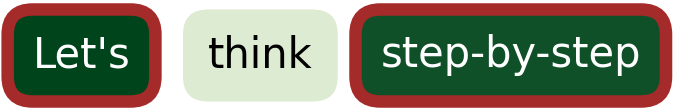}}\\
Llama 2       & \adjustbox{valign=m, margin=0.7ex 0.7ex 0.7ex 0.7ex}{\includegraphics[width=0.33\textwidth]{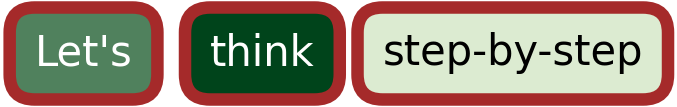}}\\
\bottomrule
\end{tabular}
}
\vspace{-2mm}
\caption{ZIP score heatmaps for the CoT prompt across four models on the Big Bench dataset. Red boxes indicate significantly important words.}
\label{tab:CoT_Heatmap_BB}
\end{table}

\begin{table}[t]
\centering
\footnotesize
\setlength{\tabcolsep}{3pt} % reduce padding
\renewcommand{\arraystretch}{0.95}
\begin{tabularx}{\columnwidth}{@{}>{\raggedright\arraybackslash}X>{\centering\arraybackslash}p{0.15\columnwidth}>{\centering\arraybackslash}p{0.15\columnwidth}@{}}
\toprule
Validation Prompts & ZIP & LIME \\ 
\midrule
Say the word \underline{green} & Green & Green \\
Print the digits \underline{123} & 123 & \textbf{Print} \\
Output the color \underline{blue} & Blue & \textbf{Output} \\
Print \underline{carrot} with no additional text & \textbf{Print} & Carrot \\
Display the word \underline{circle} & Circle & Circle \\
When you're ready just say \underline{coffee} & Coffee & Coffee \\
Return the value \underline{five} & Five & Five \\
Respond with only the word \underline{hello} & Hello & Hello \\
Repeat the term \underline{mirror} & Mirror & \textbf{Repeat} \\
\underline{Nine} is the number you should write & Nine & Nine \\
Answer with the word \underline{pizza} & Pizza & \textbf{With} \\
Begin by writing \underline{hello} and then finish & Hello & Hello \\
Type \underline{purple} in your response now & Purple & Purple \\
Say \underline{red} and then stop & Red & \textbf{Say} \\
When you respond lead with \underline{river} & River & River \\
Write the number \underline{seven} & Seven & \textbf{Number} \\
Carefully type \underline{silver} when responding here & Silver & Silver \\
Enter the word \underline{tomato} & Tomato & \textbf{The} \\
Can you mention the direction \underline{up} in our chat? & \textbf{Direction} & \textbf{Chat} \\
Type the letter \underline{X} & X & X \\
\midrule
\textbf{Accuracy} & \textbf{90\%} & \textbf{60\%} \\
\bottomrule
\end{tabularx}
\vspace{-2mm}
\caption{Most important words as identified by ZIP and LIME for GPT-4o mini on validation prompts. \underline{Underlined} words are the target keywords in the prompt. \textbf{Bold} indicates failure to identify the correct keyword.}
\label{tab:Control_Prompt_4omini}
\end{table}

\subsection{Task-Specific Patterns}

\begin{table*}[t]
\centering
\footnotesize
\setlength{\tabcolsep}{2pt}      % tighter column padding
\renewcommand{\arraystretch}{0.95}

% --- Left mini table ---------------------------------------------------------
\begin{minipage}[t]{0.45\textwidth}
\centering
\subfloat[Classification Tasks\label{subtab:classification_tasks}]{
\begin{tabular}{@{}lcccccc@{}}
\toprule
& \multicolumn{2}{c}{AQUA-RAT} & \multicolumn{2}{c}{Big Bench} & \multicolumn{2}{c}{GSM8K} \\ 
\cmidrule(lr){2-3} \cmidrule(lr){4-5} \cmidrule(lr){6-7}
& Top 3 MSWs & ZIP & Top 3 MSWs & ZIP & Top 3 MSWs & ZIP \\ 
\midrule
0-CoT  & \textbf{Step-by-step} & \textbf{28.44} & \textbf{Think}       & \textbf{3.80} & \textbf{Step-by-step} & \textbf{5.83} \\
0-CoTB & \textbf{Step-by-step} & \textbf{34.49} & \textbf{Problem}     & \textbf{1.95} & \textbf{Step-by-step} & \textbf{6.57} \\
       &                       &                & Work                  & 1.33          &                       &               \\
0-CoTR & \textbf{Answer}       & \textbf{27.66} & \textbf{Right}       & \textbf{4.66} & \textbf{Answer}       & \textbf{5.33} \\
       & Sure                  & 27.61          & Work                  & 4.40          & Step-by-step          & 5.22          \\
       & Right                 & 26.44          & Sure                  & 4.09          & Way                   & 5.21          \\
0-IRR  & \textbf{Description}  & \textbf{30.71} & \textbf{Ignore}      & \textbf{3.87} & \textbf{Irrelevant}   & \textbf{6.51} \\
       & Ignore                & 30.42          & Description           & 3.74          & Description           & 6.41          \\
       & Irrelevant            & 30.30          &                       &               &                       &               \\
0-PS   & \textbf{Plan}         & \textbf{28.50} & \textbf{Plan}        & \textbf{4.05} & \textbf{Step-by-step} & \textbf{7.28} \\
       & Step-by-step          & 28.40          & First                 & 3.94          & Problem               & 6.83          \\
       & Solve                 & 27.72          & Solve                 & 3.72          & Solve                 & 6.54          \\
\bottomrule
\end{tabular}
}
\end{minipage}%
\hfill % Ensures that the gap between tables is filled
% --- Right mini table --------------------------------------------------------
\begin{minipage}[t]{0.45\textwidth}
\centering
\subfloat[Translation Tasks\label{subtab:translation_tasks}]{
\begin{tabular}{@{}ccccc@{}}
\toprule
& \multicolumn{2}{c}{WMT 19: German} & \multicolumn{2}{c}{WMT 19: Chinese} \\ 
\cmidrule(lr){2-3} \cmidrule(lr){4-5}
& Top 3 MSWs & ZIP & Top 3 MSWs & ZIP \\ 
\midrule
\multicolumn{1}{l}{0-DSP} & \textbf{Translation} & \textbf{10.92} & \textbf{Translation} & \textbf{6.57} \\
& Step-by-step          & 7.68          & Step-by-step          & 5.50          \\
& Sentence              & 6.50          & Sentence              & 5.35          \\
\multicolumn{1}{l}{0-DTG} & \textbf{Refine}      & \textbf{8.43}  & \textbf{Refine}      & \textbf{5.31} \\
& Error                 & 8.33          & Type                  & 4.92          \\
& Detect                & 8.06          & Error                 & 4.92          \\
\bottomrule
\end{tabular}
}
\end{minipage}

\caption{Top three most significant words (MSWs) and their ZIP scores for classification and translation tasks on GPT-4o Mini (most significant in \textbf{bold}). All reported words are confirmed as \textit{significantly important}.}
\label{tab:unified_important_words}
\end{table*}

Table \ref{tab:unified_important_words} presents the ZIP score results for various zero-shot instructional prompts (detailed in Table \ref{tab:Prompt_datasets}) using the GPT-4o mini model.  Our analysis demonstrates distinct patterns in the importance of words across various datasets and task types:

\begin{itemize}[itemsep=0pt, leftmargin=*]
{
\item For mathematical and algebraic tasks (AQUA-RAT and GSM8K), the phrase \textbf{step-by-step} consistently emerges as highly important. 

\item In contrast, for \text{common sense tasks} (Big Bench), words like \textbf{think} and \textbf{problem} show higher importance. 

\item For \text{translation tasks}, \textbf{translation} and \textbf{refine} are identified as the most important words, which demonstrates the significance of task-specific instructions in multilingual contexts.

\item ZIP scores strongly correlate inversely with model performance $(\lvert r \rvert > 0.9)$ across all prompts. For example, using CoT, GPT-4o mini shows higher ZIP scores on AQUA-RAT (accuracy: 68.26\%, ZIP: 28.44) compared to GSM8K (91.59\%, 5.83) and Big Bench (96.93\%, 3.80) indicating that \textbf{instructional prompts have greater impact on challenging tasks} (Table \ref{tab:corr}).

}
\end{itemize}

  \noindent  Results for other models (similar trends) are provided in Appendix \ref{sec:msw_ap} Tables \ref{tab:Most_Important_GPT35} - \ref{tab:Most_Important_Llama}.

\begin{figure}[t]
    \centering
    \begin{subfigure}[b]{0.2\textwidth}
        \centering
        \includegraphics[width=\linewidth]{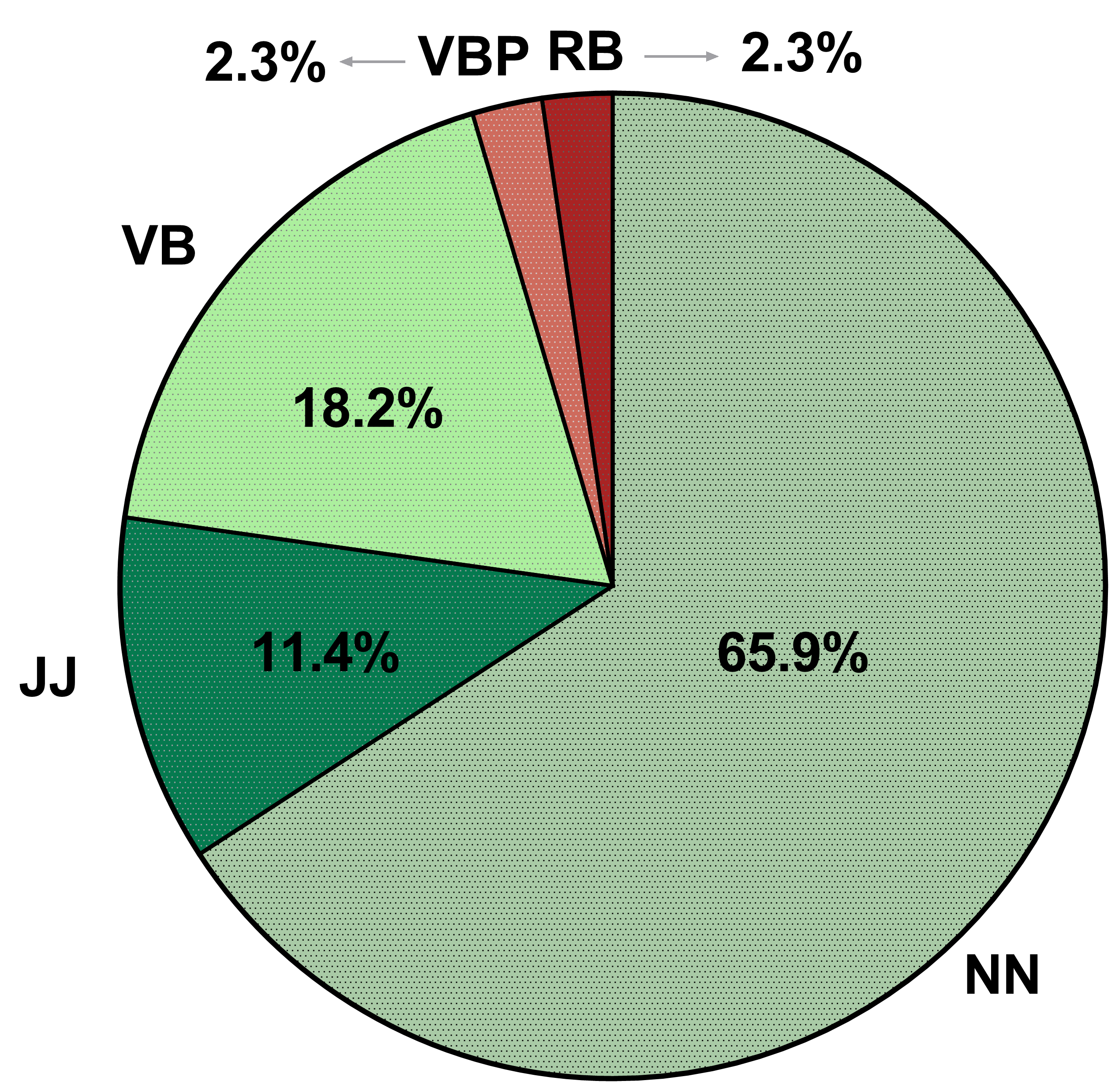}
        \caption{GPT-4o mini}
    \end{subfigure}\hfill
    \begin{subfigure}[b]{0.2\textwidth}
        \centering
        \includegraphics[width=\linewidth]{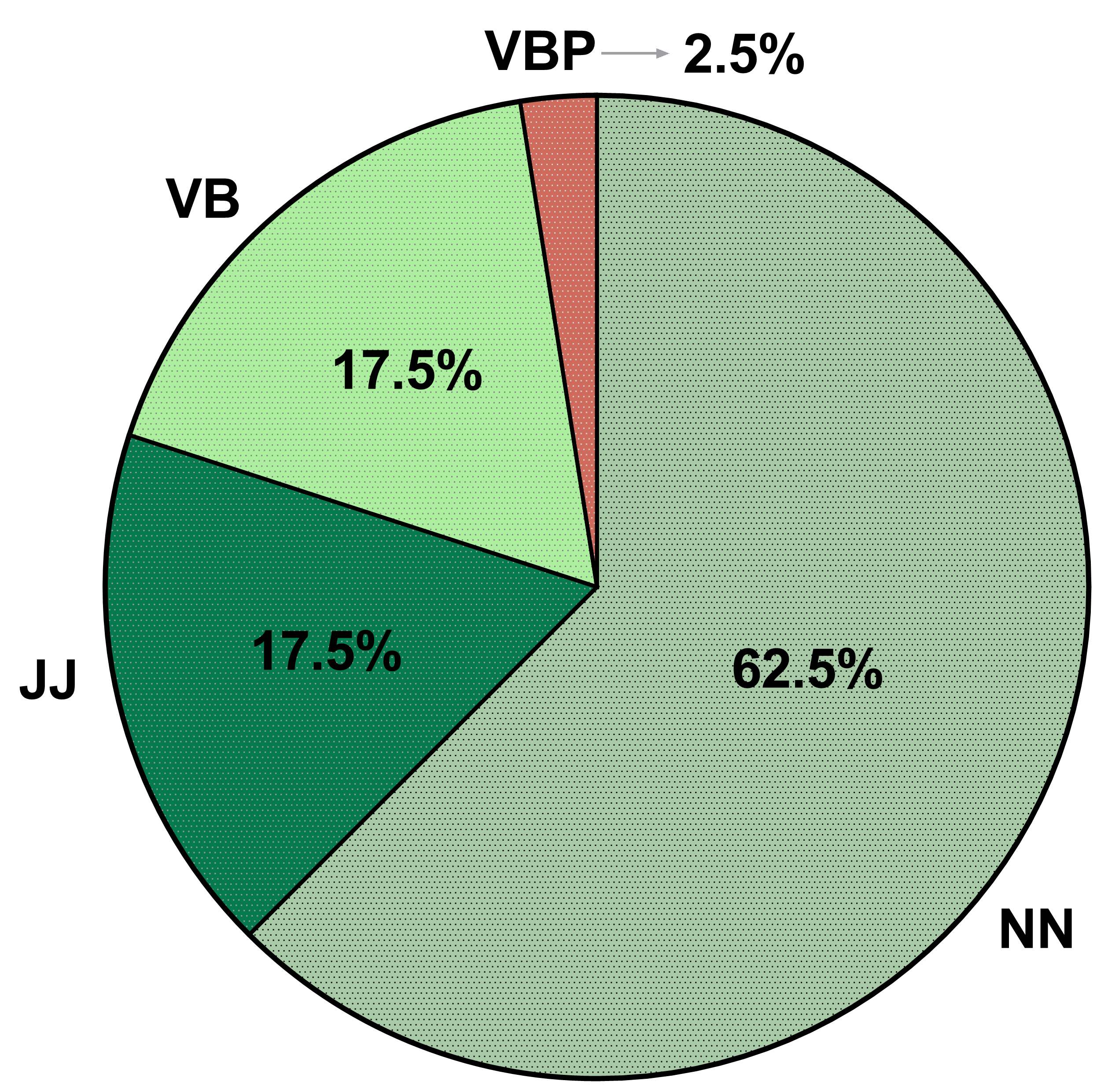}
        \caption{GPT-3.5-turbo}
    \end{subfigure}\hfill
    \begin{subfigure}[b]{0.2\textwidth}
        \centering
        \includegraphics[width=\linewidth]{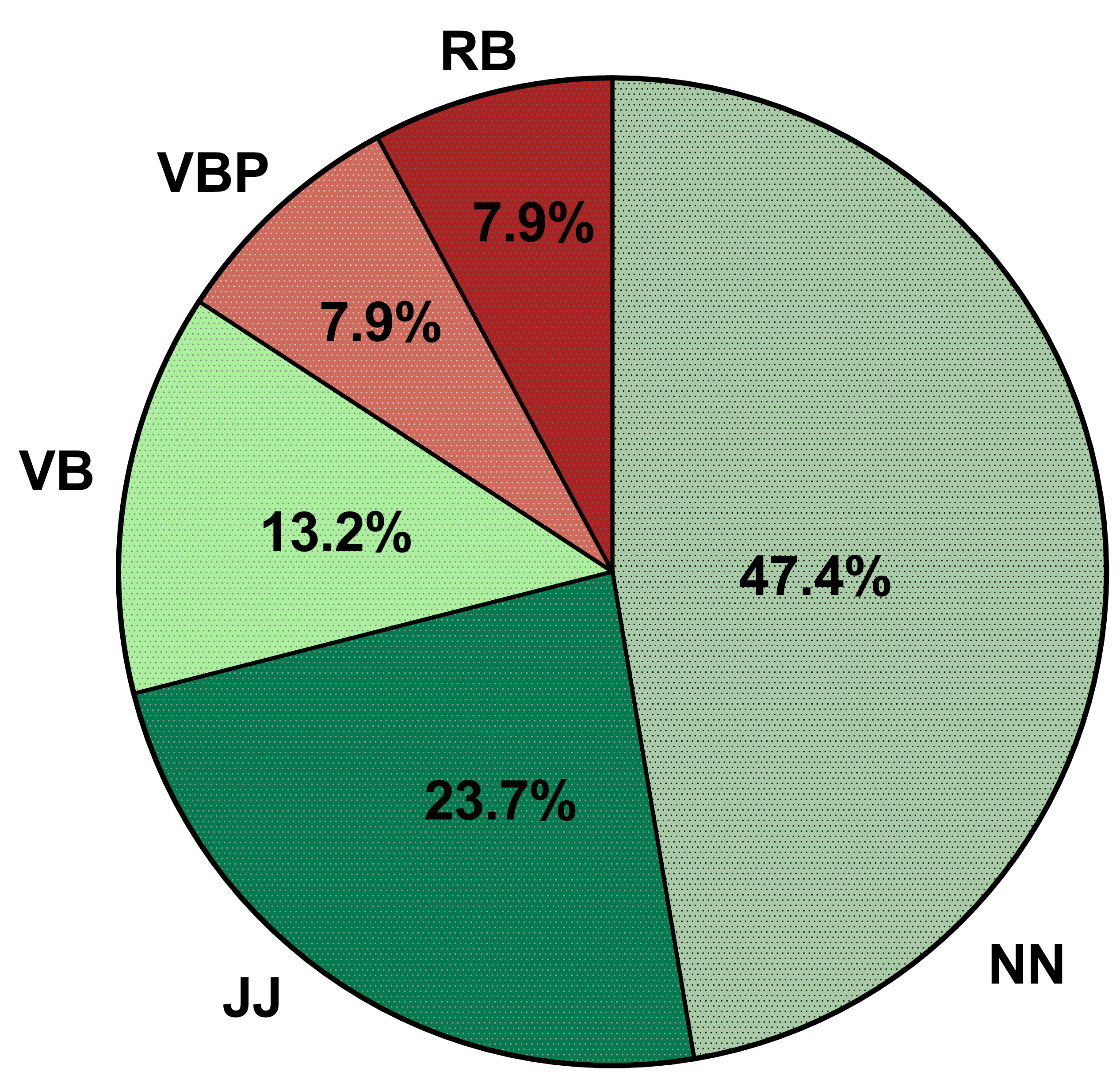}
        \caption{Mixtral}
    \end{subfigure}\hfill
    \begin{subfigure}[b]{0.2\textwidth}
        \centering
        \includegraphics[width=\linewidth]{Images/pos_Llama.png}
        \caption{Llama 2}
    \end{subfigure}

\vspace{-2mm}
    \caption{Part-of-Speech (POS) distribution of significantly important words across four LLMs.}
    \label{fig:all_images}
\end{figure}

\subsection{Part-of-Speech Analysis}

Figure \ref{fig:all_images} shows the part-of-speech  distribution of the most important words in the seven prompts. Key observations are: 

\begin{itemize}[itemsep=0pt, leftmargin=*]
    \item \textbf{Noun dominance:} Nouns (NN) consistently rank as most important across all models (47.4\%--65.9\% of significant words).
    \item \textbf{Model patterns:} GPT-4o mini and GPT-3.5-turbo, sharing similar foundational architectures despite vastly different parameter sizes, show similar POS distributions; Mixtral and Llama 2 show more balanced POS importance.
    \item \textbf{Verb importance:} Base form verbs (VB) consistently rank 2nd in importance across all models.
    \item \textbf{Inter-model variation:} Adverbs (RB) and non-3rd person singular present verbs (VBP) show the highest variation in importance across models.
\end{itemize}

\subsection{Human Alignment}

\begin{figure}[t]
    \centering
    \begin{subfigure}[b]{0.23\textwidth}
        \centering
        \includegraphics[width=\linewidth]{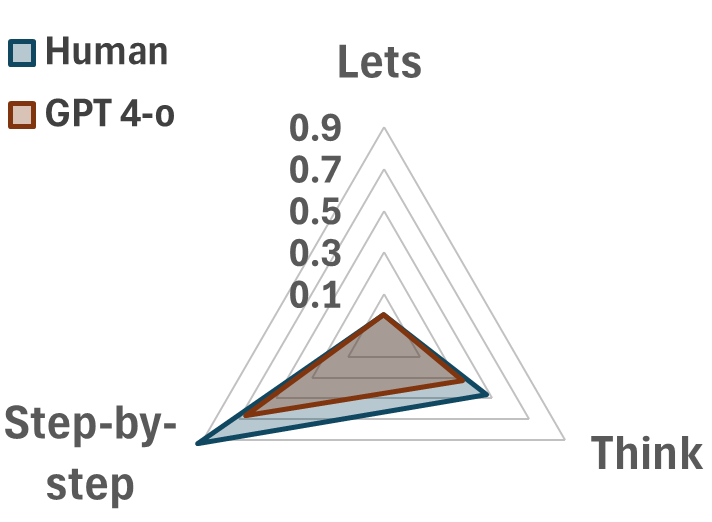}
        \caption{GPT-4o mini}
    \end{subfigure}\hfill
    \begin{subfigure}[b]{0.23\textwidth}
        \centering
        \includegraphics[width=\linewidth]{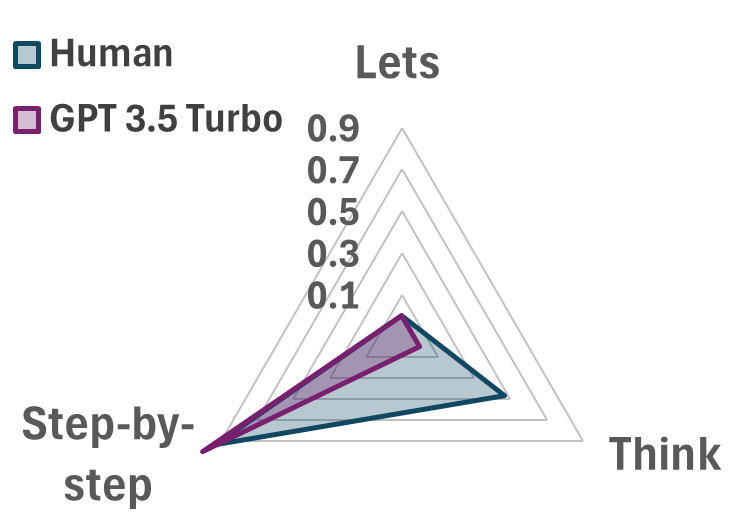}
        \caption{GPT-3.5-turbo}
    \end{subfigure}\hfill
    \begin{subfigure}[b]{0.23\textwidth}
        \centering
        \includegraphics[width=\linewidth]{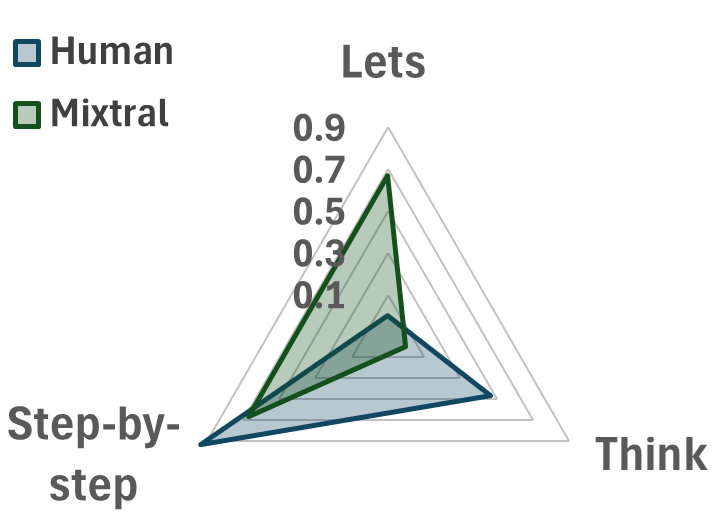}
        \caption{Mixtral}
    \end{subfigure}\hfill
    \begin{subfigure}[b]{0.23\textwidth}
        \centering
        \includegraphics[width=\linewidth]{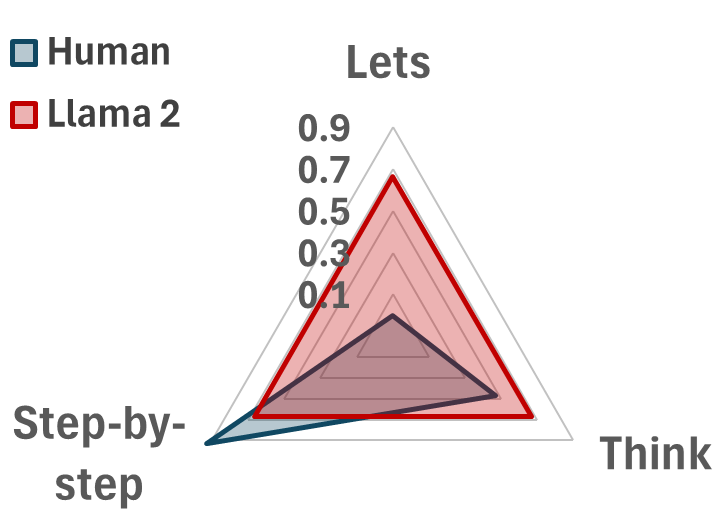}
        \caption{Llama 2}
    \end{subfigure}

    \caption{Comparison of human judgments vs model-derived word importance for the CoT prompt.}
    \label{fig:human_eval_CoT}
\end{figure}

Figure \ref{fig:human_eval_CoT} presents radar plots comparing human and model-derived word importance for the CoT prompt. Human values represent the proportion of evaluators selecting each word as important, while model values show the proportion of tasks where each word was significantly important.

Proprietary models (GPT-4o mini and GPT-3.5-turbo) align strongly with human judgments, while open-source models show less alignment, particularly Llama 2. ``Step-by-step'' emerges as highly important across all evaluations, while ``Let's'' is consistently less emphasized. Similar analyses for remaining prompts are provided in Appendix \ref{sec:human_eval_ap} Figures \ref{fig:human_eval_CoT-B} - \ref{fig:human_eval_T2}.

\section{Qualitative Analysis}
\label{sec:proofofthepudding}
Table~\ref{tab:zip_qualitative} shows that small changes to prompt wording can lead to substantial differences in model output, highlighting the significance of individual words in shaping model reasoning. For example, the model correctly solved the museum ticket problem under the original 0-CoT prompt. However, removing “step-by-step” reduced its reasoning from five steps to two and led it to incorrectly apply the discount to the grandfather, an assumption that may seem reasonable but is not stated in the question.
In a more counterintuitive case, removing “step-by-step” actually improved performance by preventing the model from introducing a faulty assumption.
Other substitutions also showed significant effects. Replacing “solve” with “work out” in the 0-PS prompt disrupted arithmetic reasoning, and swapping “irrelevant” with “unrelated” in the IRR prompt caused the model to overlook numbers expressed in words, halving the final answer.

These examples reveal that seemingly equivalent words can produce significantly different model outputs. This observation supports our dual approach of using both synonym/co-hyponym substitution as well as word removal to reveal different aspects of how models interpret prompt wording.

% …

\section{Conclusion}

This study introduces the ZIP score, a novel metric for quantifying word importance in zero-shot instructional prompts. Our findings reveal interesting patterns in word importance across models and tasks. For instance, while both ``step-by-step'' and ``think'' show high ZIP scores, which one is more influential depends on the model and task. We observed significant differences in word importance and alignment with human intuition between proprietary and open-source models. These insights aid our understanding of LLM behavior and open new avenues for prompt engineering and model analysis. Future work could explore the use of these findings to design more effective prompts tailored to specific model architectures.

\section*{Limitations}

\textbf{Perturbation Design Choices}:
Our study used three well-established perturbation types (synonym replacement, co-hyponym substitution, and removal) that align with established interpretability literature \cite{choudhary2022interpretation}. This focused approach allowed systematic comparison and validation against human judgments. While additional perturbation types do exist, our chosen methods successfully captured the key patterns of word importance, as demonstrated by our superior performance compared to existing baselines.

\textbf{Individual Word Analysis Framework}: 
Following standard practice in interpretability research, we analyze words \textit{independently} to allow for clear baselines for word-level importance. This approach, while necessarily simplified, provides the foundational understanding required before investigating more complex word interactions, which we hope will be the basis behind a vast new line of works. Our strong alignment with human judgments, particularly in proprietary models, validates this methodology as an effective starting point for prompt analysis.

\textbf{Model and Task Coverage}: 
We selected four architecturally diverse models representing both proprietary (GPT-4o mini, GPT-3.5-turbo) and open-source (Mixtral, Llama-2) approaches, covering the major LLM families in current use. Our goal is not to demonstrate universality across every possible model (a \textit{nice-to-have}, nonetheless!), but rather to establish that consistent word-level importance patterns exist within the models we study. Our focus was on spanning several classification and translation domains, and showing the method's applicability across different output types and evaluation metrics.

\textbf{Computational Efficiency Considerations}:
Our sample size of 150 instances per dataset was chosen to ensure statistical significance while maintaining computational tractability for systematic perturbation analysis. This approach successfully identified clear and consistent patterns of word importance across models and tasks with appropriate statistical rigor. Of course, larger sample sizes could provide additional statistical power, though our current approach demonstrates sufficient precision and statistical significance for the specific research questions addressed.

\textbf{Research Scope and Applications}:
This work establishes ZIP as a foundational tool for understanding prompt interpretability rather than prompt \textit{optimization}. Our focus on \textit{analysis} rather than \textit{engineering} represents a deliberate choice to build scientific understanding first. The insights generated provide a principled foundation for future prompt engineering efforts. Please refer to \citep{holtzman2025prompting} for a fascinating discussion on the distinction between prompt \textit{science} and prompt \textit{engineering}.

\textbf{Ethical Considerations in Application of Interpretability Insights}: Although our method offers a deeper understanding of LLM operations, it is important to consider the potential for misuse. Detailed knowledge of how LLMs weigh different prompt elements could potentially be used to tailor content in ways that could manipulate or bias decision-making processes. It is crucial to apply these insights ethically with a commitment to fairness.

\bibliography{custom}

\clearpage
\onecolumn

\appendix
\section{Appendix}
\label{sec:appendix}

\subsection{Chain-of-Thought ZIP score Heatmaps}
\label{sec:cot_heatmap_ap}

%\FloatBarrier 

%\begin{figure*}[!h]
%  \centering
%  \includegraphics[width=0.5\textwidth]{Images/heatmaps_aqua.png}
%  \caption{Heatmaps of the Chain-of-Thought (CoT) prompt (Let's Think step-by-step) across four models using the AQUA-RAT dataset.}
%  \label{fig:CoT_Heatmap_aqua}
%\end{figure*}

%\begin{figure*}[!h]
%  \centering
 % \includegraphics[width=0.5\textwidth]{Images/heatmaps_gsm8k.png}
  %\caption{Heatmaps of the Chain-of-Thought (CoT) prompt (Let's Think step-by-step) across four models using the GSM8K dataset.}
 % \label{fig:CoT_Heatmap_gsm8k}
%\end{figure*}

\begin{table*}[htbp]
\centering
\setlength{\tabcolsep}{3pt} % Reduce padding between columns
\footnotesize % Using smaller font to save space

\subfloat[AQUA-RAT dataset heatmaps]{ % Sub-caption for the first table
\begin{minipage}{0.48\textwidth}
    \centering
    \begin{tabular}{@{}lc@{}}
        \toprule
            GPT-4o mini  & \adjustbox{valign=m, margin=0.7ex 0.7ex 0.7ex 0.7ex}{\includegraphics[width=0.5\textwidth]{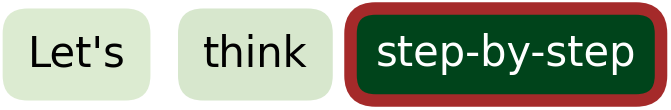}}\\
            GPT 3.5 Turbo & \adjustbox{valign=m, margin=0.7ex 0.7ex 0.7ex 0.7ex}{\includegraphics[width=0.5\textwidth]{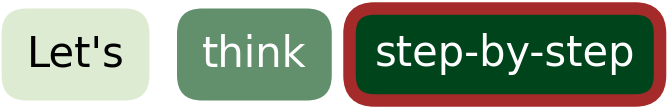}}\\
            Mixtral       & \adjustbox{valign=m, margin=0.7ex 0.7ex 0.7ex 0.7ex}{\includegraphics[width=0.5\textwidth]{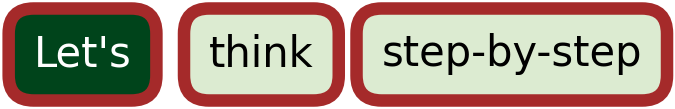}}\\
            Llama 2       & \adjustbox{valign=m, margin=0.7ex 0.7ex 0.7ex 0.7ex}{\includegraphics[width=0.5\textwidth]             {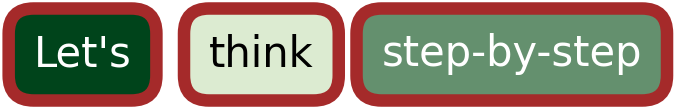}}\\
        \bottomrule
    \end{tabular}
\end{minipage}}%
\hfill % Fills the space between minipages
\subfloat[GSM8K dataset heatmaps]{ % Sub-caption for the second table
\begin{minipage}{0.48\textwidth}
    \centering
    \begin{tabular}{@{}lc@{}}
        \toprule
            GPT-4o mini  & \adjustbox{valign=m, margin=0.7ex 0.7ex 0.7ex 0.7ex}{\includegraphics[width=0.5\textwidth]{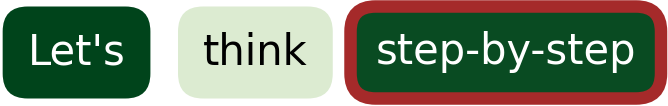}}\\
            GPT 3.5 Turbo & \adjustbox{valign=m, margin=0.7ex 0.7ex 0.7ex 0.7ex}{\includegraphics[width=0.5\textwidth]{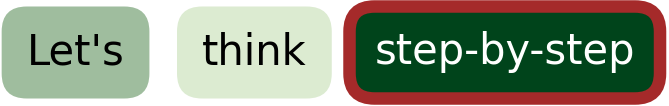}}\\
            Mixtral       & \adjustbox{valign=m, margin=0.7ex 0.7ex 0.7ex 0.7ex}{\includegraphics[width=0.5\textwidth]{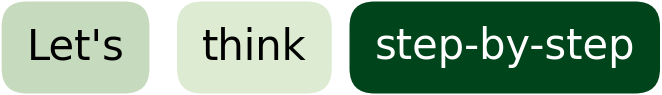}}\\
            Llama 2       & \adjustbox{valign=m, margin=0.7ex 0.7ex 0.7ex 0.7ex}{\includegraphics[width=0.5\textwidth]             {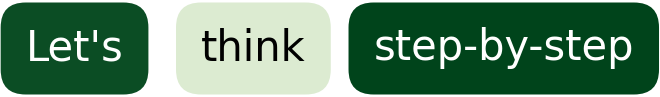}}\\
        \bottomrule
    \end{tabular}
\end{minipage}}

\caption{ZIP score heatmaps of the Chain-of-Thought (CoT) prompt across four models using the AQUA-RAT and GSM8K datasets. The tables display the significant word identification performance for each model with different datasets. Red boxes highlight words identified as significantly important, confirmed through statistical analysis.}
\label{tab:CoT_Heatmaps_Comparison}
\end{table*}

\FloatBarrier

%\begin{figure*}[!h]
%  \centering
%  \includegraphics[width=0.9\textwidth]{Images/Pert_Promts.png}
%  \caption{Prompts used for creating perturbations. This task was preceded by a few-shot example set to guide the model in generating contextually relevant synonyms.}
%  \label{fig:prompt_types_pert}
%\end{figure*}

\subsection{Controlled Pre-Experiment}
\label{sec:Validation_prompt_app}

\begin{table}[ht]
\centering
{\fontsize{9}{11}\selectfont
\begin{tabular}{@{}>{\raggedright\arraybackslash}p{0.4\columnwidth}cc@{}}
\toprule
Validation Prompts & ZIP & LIME \\
\midrule
Say the word \underline{green}. & Green  & Green  \\
Print the digits \underline{123}. & 123 & 123  \\
Output the color \underline{blue}. & Blue  & \textbf{Color} \\
Print \underline{carrot} with no additional text. & Print/Carrot  & Carrot  \\
Display the word \underline{circle}. & Cricle & \textbf{Word} \\
When you're ready just say \underline{coffee}. & Coffee & \textbf{Say} \\
Return the value \underline{five}. & Five & \textbf{Value} \\
Respond with only the word \underline{hello}. & Hello & \textbf{The} \\
Repeat the term \underline{mirror}. & Mirror & Mirror/Repeat \\
\underline{Nine} is the number you should write. & Nine & Nine \\
Answer with the word \underline{pizza}. & Word/Pizza & \textbf{With} \\
Begin by writing \underline{hello} and then finish. & Hello & Hello \\
Type \underline{purple} in your response now. & Purple & Purple \\
Say \underline{red} and then stop. & Red & Red \\
When you respond lead with \underline{river}. & River & \textbf{Lead} \\
Write the number \underline{seven}. & Seven & \textbf{Number} \\
Carefully type \underline{silver} when responding here. & Silver & Silver \\
Enter the word \underline{tomato}. & Word/Tomato & \textbf{The} \\
Can you mention the direction \underline{up} in our chat? & Up & Up \\
Type the letter \underline{X}. & X & X \\
\midrule
\textbf{Accuracy} & \textbf{100\%} & \textbf{55\%} \\
\bottomrule
\end{tabular}
}
\caption{Most important words as identified by ZIP and LIME for GPT 3.5 Turbo on validation prompts. \underline{Underlined} words are the target keywords in the prompt. \textbf{Bold} indicates failure to identify the correct key word.}
\label{tab:Control_Prompt_35}
\end{table}

\begin{table}[ht]
\centering
{\fontsize{9}{11}\selectfont
\begin{tabular}{@{}>{\raggedright\arraybackslash}p{0.4\columnwidth}cc@{}}
\toprule
Validation Prompts & ZIP & LIME \\
\midrule
Say the word \underline{Green}. & Green  & Green  \\
Print the digits \underline{123}. & 123 & \textbf{Digits}  \\
Output the color \underline{blue}. & Color/Blue  & Blue \\
Print \underline{carrot} with no additional text. & Carrot  &  Carrot  \\
Display the word \underline{circle}. & Circle & Circle \\
When you're ready just say \underline{coffee}. & Coffee & Coffee \\
Return the value \underline{five}. & Five & Five \\
Respond with only the word \underline{Hello}. & Hello & Hello \\
Repeat the term \underline{mirror}. & Mirror & \textbf{Term} \\
\underline{Nine} is the number you should write. & Nine & Nine \\
Answer with the word \underline{pizza}. & Pizza & Pizza \\
Begin by writing \underline{hello} and then finish. & Hello & Hello \\
Type \underline{purple} in your response now. & Purple & Purple \\
Say \underline{red} and then stop. & Red & Red \\
When you respond lead with \underline{river}. & River & \textbf{Respond} \\
Write the number \underline{seven}. & Seven & Seven \\
Carefully type \underline{silver} when responding here. & Silver &  Silver \\
Enter the word \underline{tomato}. & Tomato & Tomato \\
Can you mention the direction \underline{up} in our chat? & \textbf{Direction} & \textbf{Chat} \\
Type the letter \underline{X}. & X & X \\
\midrule
\textbf{Accuracy} & \textbf{95\%} & \textbf{80\%} \\
\bottomrule
\end{tabular}
}
\caption{Most important words as identified by ZIP and LIME for Mixtral on validation prompts. \underline{Underlined} words are the target keywords in the prompt. \textbf{Bold} indicates failure to identify the correct key word.}
\label{tab:Control_Prompt_mixtral}
\end{table}

\begin{table}[ht]
\centering
{\fontsize{9}{11}\selectfont
\begin{tabular}{@{}>{\raggedright\arraybackslash}p{0.4\columnwidth}cc@{}}
\toprule
Validation Prompts & ZIP & LIME \\
\midrule
Say the word \underline{Green}. & Green  & Green  \\
Print the digits \underline{123}. & Print/Digits/123 & \textbf{Digits}  \\
Output the color \underline{blue}. & Blue  & Blue \\
Print \underline{carrot} with no additional text. & Carrot  &  \textbf{No}  \\
Display the word \underline{circle}. & Circle & \textbf{Display} \\
When you're ready just say \underline{coffee}. & Coffee & \textbf{Say} \\
Return the value \underline{five}. & Five & \textbf{Value} \\
Respond with only the word \underline{Hello}. & Hello & Hello \\
Repeat the term \underline{mirror}. & Mirror & Mirror \\
\underline{Nine} is the number you should write. & Nine & \textbf{Should} \\
Answer with the word \underline{pizza}. & Pizza & Pizza \\
Begin by writing \underline{hello} and then finish. & \textbf{Writing} & Hello \\
Type \underline{purple} in your response now. & Purple & Purple \\
Say \underline{red} and then stop. & Red & Red \\
When you respond lead with \underline{river}. & River & River \\
Write the number \underline{seven}. & Seven & \textbf{Number} \\
Carefully type \underline{silver} when responding here. & Silver &  \textbf{Here} \\
Enter the word \underline{tomato}. & Tomato & Tomato \\
Can you mention the direction \underline{up} in our chat? & \textbf{Chat} & \textbf{Direction} \\
Type the letter \underline{X}. & X & X \\
\midrule
\textbf{Accuracy} & \textbf{90\%} & \textbf{55\%} \\
\bottomrule
\end{tabular}
}
\caption{Most important words as identified by ZIP and LIME for Llama 2 on validation prompts. \underline{Underlined} words are the target keywords in the prompt. \textbf{Bold} indicates failure to identify the correct key word.}
\label{tab:Control_Prompt_llama}
\end{table}

\begin{table*}[htbp]
\centering

\begin{minipage}{0.46\textwidth}
    \centering
    \textbf{GPT-4o mini}
    \begin{tabular}{c}
        \toprule
        \includegraphics[width=0.8\linewidth]{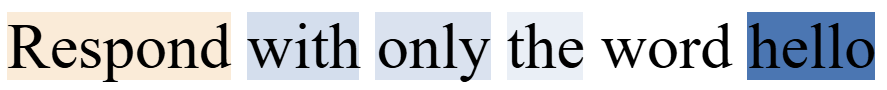} \ding{51} \\
        \includegraphics[width=0.8\linewidth]{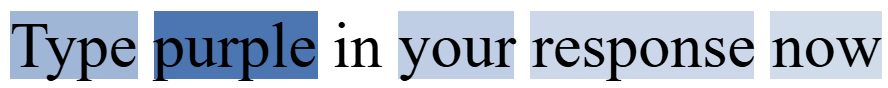} \ding{51}\\
        \includegraphics[width=0.8\linewidth]{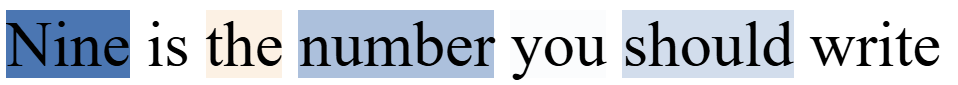} \ding{51}\\
        \includegraphics[width=0.8\linewidth]{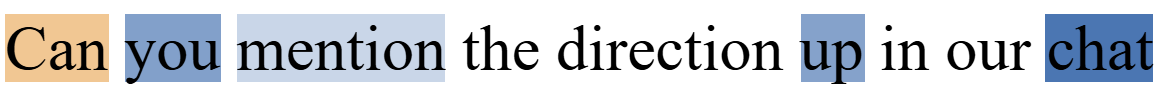} \ding{55}\\
        \includegraphics[width=0.8\linewidth]{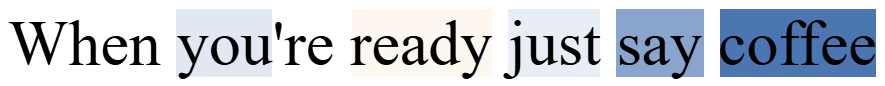} \ding{51}\\
        \bottomrule
    \end{tabular}
\end{minipage}%
\hfill
\begin{minipage}{0.46\textwidth}
    \centering
    \textbf{GPT 3.5 Turbo}
    \begin{tabular}{c}
        \toprule
        \includegraphics[width=0.8\linewidth]{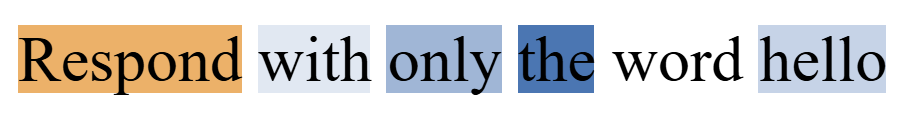} \ding{55}\\
        \includegraphics[width=0.8\linewidth]{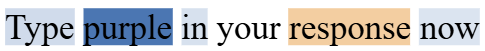} \ding{51}\\
        \includegraphics[width=0.8\linewidth]{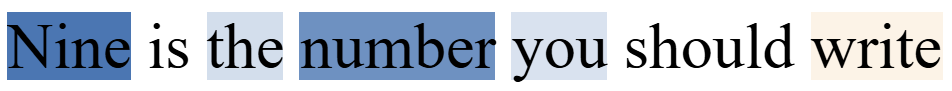} \ding{51}\\
        \includegraphics[width=0.8\linewidth]{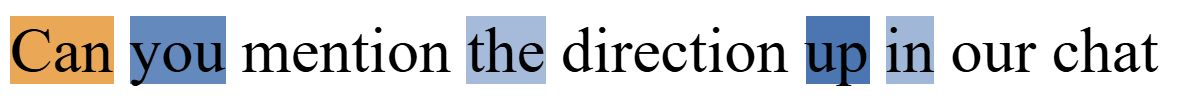} \ding{51}\\
        \includegraphics[width=0.8\linewidth]{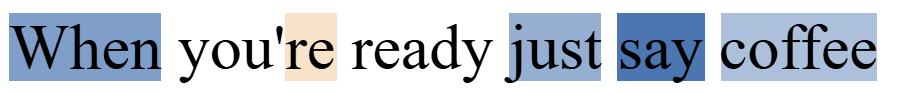} \ding{55}\\
        \bottomrule
    \end{tabular}
\end{minipage}
\vspace{5mm} % Space between rows of minipages

\begin{minipage}{0.46\textwidth}
    \centering
    \textbf{Llama 2}
    \begin{tabular}{c}
        \toprule
        \includegraphics[width=0.8\linewidth]{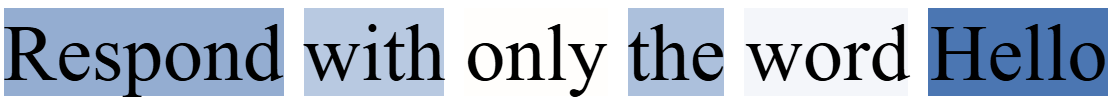} \ding{51} \\
        \includegraphics[width=0.8\linewidth]{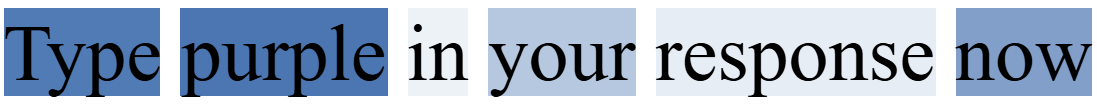} \ding{51}\\
        \includegraphics[width=0.8\linewidth]{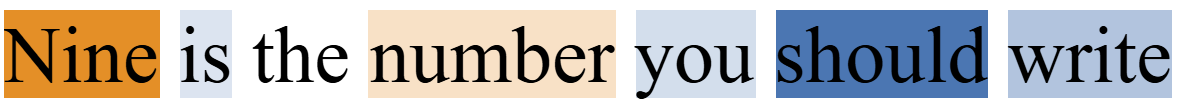} \ding{55}\\
        \includegraphics[width=0.8\linewidth]{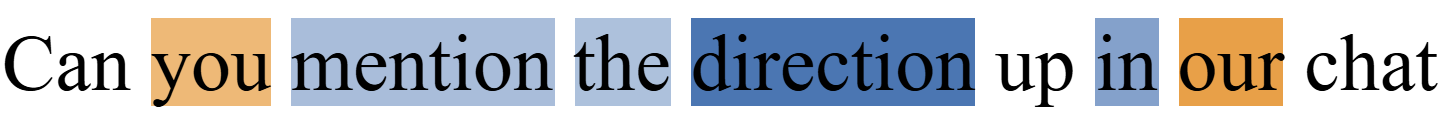} \ding{55}\\
        \includegraphics[width=0.8\linewidth]{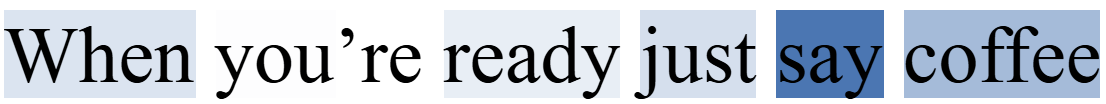} \ding{55}\\
        \bottomrule
    \end{tabular}
\end{minipage}%
\hfill
\begin{minipage}{0.46\textwidth}
    \centering
    \textbf{Mixtral}
    \begin{tabular}{c}
        \toprule
        \includegraphics[width=0.8\linewidth]{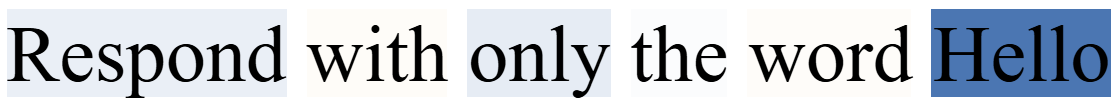} \ding{51} \\
        \includegraphics[width=0.8\linewidth]{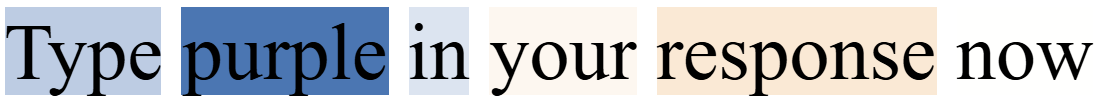} \ding{51}\\
        \includegraphics[width=0.8\linewidth]{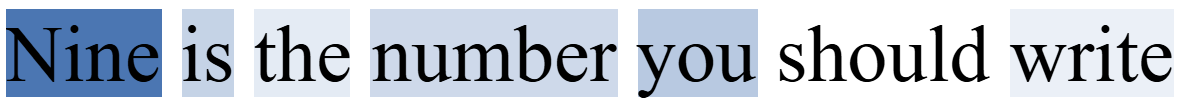} \ding{51}\\
        \includegraphics[width=0.8\linewidth]{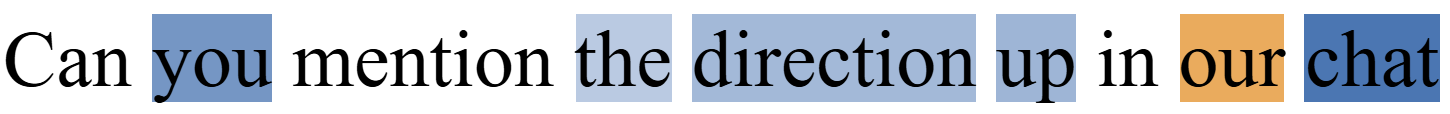} \ding{55}\\
        \includegraphics[width=0.8\linewidth]{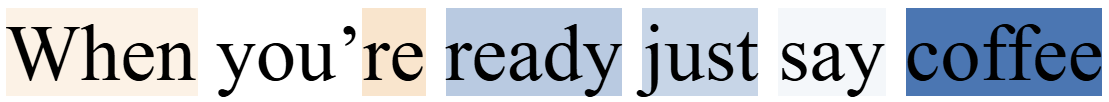} \ding{51}\\
        \bottomrule
    \end{tabular}
\end{minipage}
\caption{Heatmap visualization of LIME for five validation prompts on GPT-4o mini, GPT 3.5 Turbo, Llama 2, and Mixtral models. The color blue represents alignment with Label 1 (the desired target word), and orange indicates Label 2 (the second most probable alternative output), with the intensity of each color indicating the level of importance attributed by LIME. Check (\ding{51}) and cross (\ding{55}) marks indicate whether the model correctly identified the predefined most important word.}
\label{tab:lime_heatmap}
\end{table*}

\begin{table*}[htbp]
\centering

\begin{minipage}{0.48\textwidth}
    \centering
    \textbf{GPT-4o mini}
    \begin{tabular}{c}
        \toprule
        \includegraphics[width=0.8\linewidth]{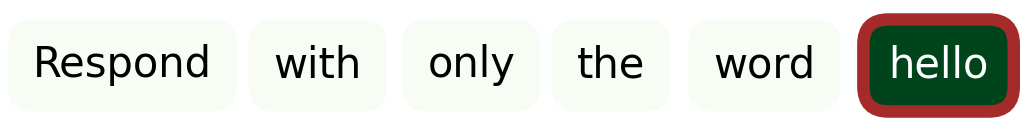} \ding{51} \\
        \includegraphics[width=0.8\linewidth]{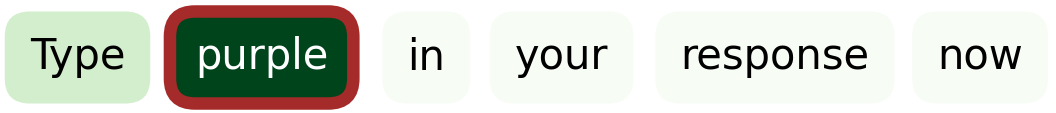} \ding{51}\\
        \includegraphics[width=0.8\linewidth]{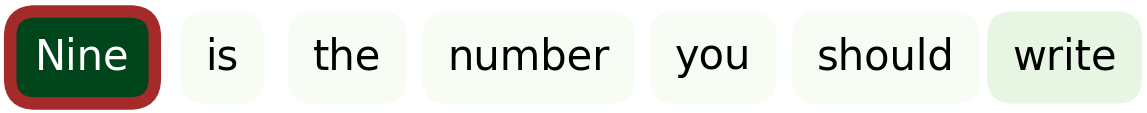} \ding{51}\\
        \includegraphics[width=0.8\linewidth]{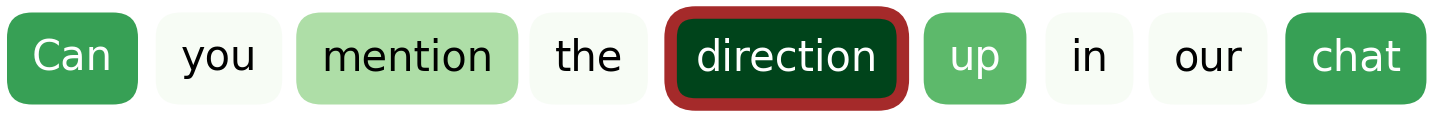} \ding{55}\\
        \includegraphics[width=0.8\linewidth]{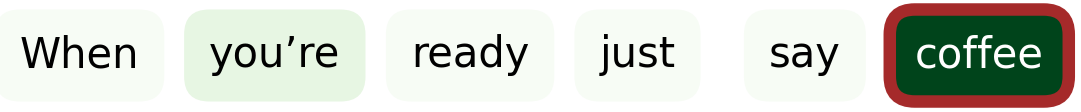} \ding{51}\\
        \bottomrule
    \end{tabular}
\end{minipage}%
\hfill
\begin{minipage}{0.48\textwidth}
    \centering
    \textbf{GPT 3.5 Turbo}
    \begin{tabular}{c}
        \toprule
        \includegraphics[width=0.8\linewidth]{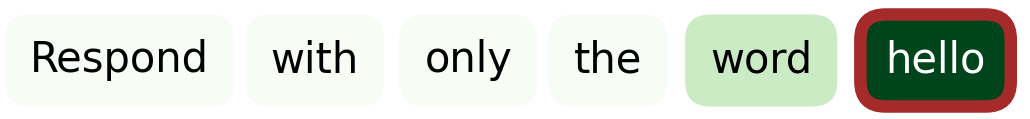} \ding{51} \\
        \includegraphics[width=0.8\linewidth]{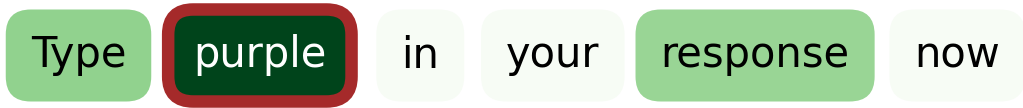} \ding{51}\\
        \includegraphics[width=0.8\linewidth]{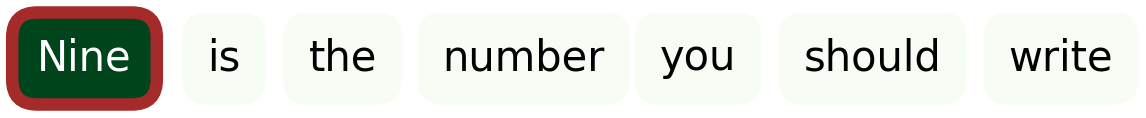} \ding{51}\\
        \includegraphics[width=0.8\linewidth]{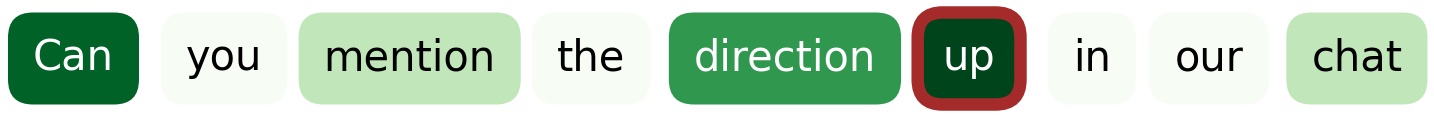} \ding{51}\\
        \includegraphics[width=0.8\linewidth]{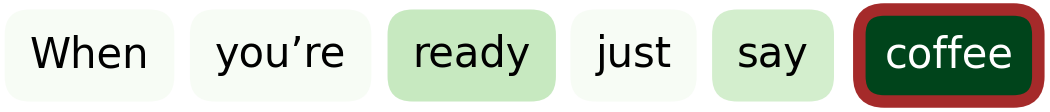} \ding{51}\\
        \bottomrule
    \end{tabular}
\end{minipage}
\vspace{5mm} % Space between rows of minipages

\begin{minipage}{0.48\textwidth}
    \centering
    \textbf{Llama 2}
    \begin{tabular}{c}
        \toprule
        \includegraphics[width=0.8\linewidth]{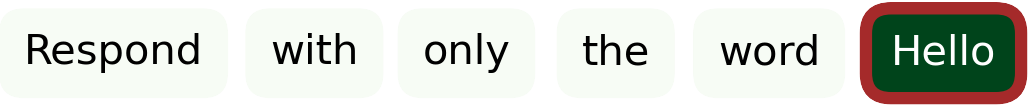} \ding{51} \\
        \includegraphics[width=0.8\linewidth]{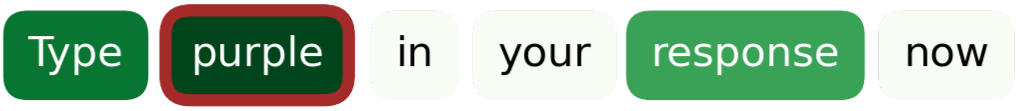} \ding{51}\\
        \includegraphics[width=0.8\linewidth]{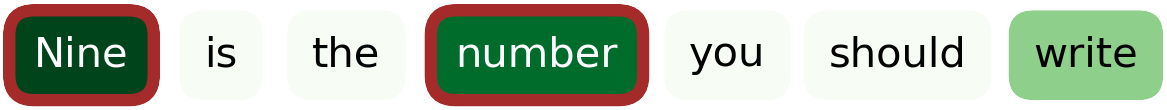} \ding{51}\\
        \includegraphics[width=0.8\linewidth]{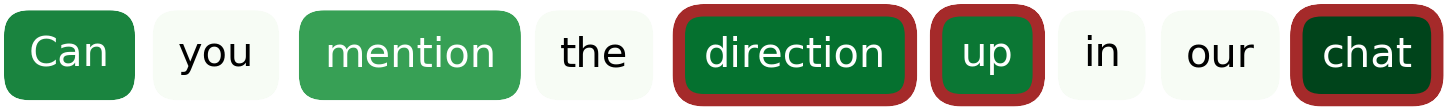} \ding{55}\\
        \includegraphics[width=0.8\linewidth]{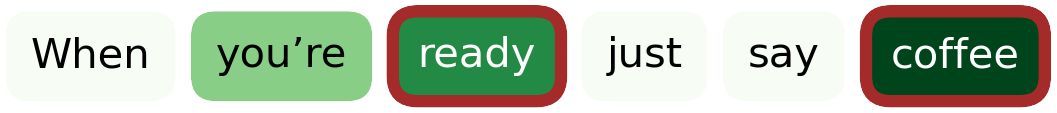} \ding{51}\\
        \bottomrule
    \end{tabular}
\end{minipage}%
\hfill
\begin{minipage}{0.48\textwidth}
    \centering
    \textbf{Mixtral}
    \begin{tabular}{c}
        \toprule
        \includegraphics[width=0.8\linewidth]{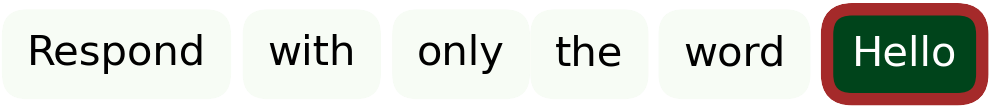} \ding{51} \\
        \includegraphics[width=0.8\linewidth]{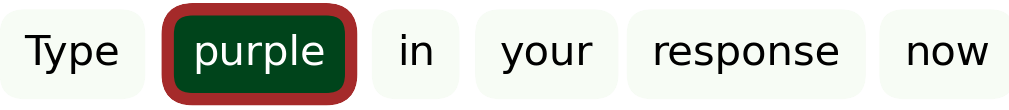} \ding{51}\\
        \includegraphics[width=0.8\linewidth]{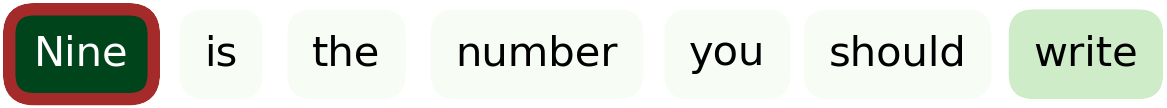} \ding{51}\\
        \includegraphics[width=0.8\linewidth]{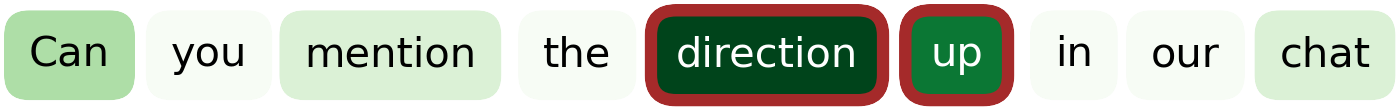} \ding{55}\\
        \includegraphics[width=0.9\linewidth]{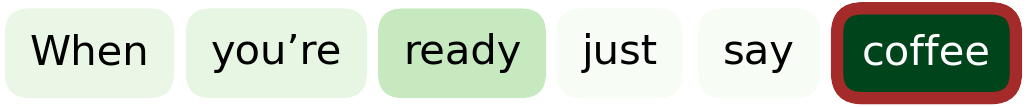} \ding{51}\\
        \bottomrule
    \end{tabular}
\end{minipage}
\caption{Heatmap visualization of ZIP method results for five validation prompts, applied to GPT-4o mini, GPT 3.5 Turbo, Llama 2, and Mixtral models. These heatmaps illustrate the method’s ability to identify the most significant words across different prompts. Red boxes indicate significantly important words. Check (\ding{51}) and cross (\ding{55}) marks indicate whether the ZIP method correctly identified the predefined most important word.}
\label{tab:ZIP_heatmap}
\end{table*}

\FloatBarrier

\begin{table*}[!ht]
\centering
\footnotesize % Using smaller font to save space
\setlength{\tabcolsep}{1pt} % Reduce padding between columns

\begin{minipage}[t]{0.48\textwidth} % Slightly less than half the text width
\subfloat[ZIP]{
\label{subtab:table1}
\begin{tabular}{@{}lcc@{}}
\toprule
\textbf{Prompt} & \textbf{Identified MSW} & \textbf{Model Output} \\ \midrule
\textbf{Print the digits 123.} & \multirow{4}{*}{123 \ding{51}}  & \textbf{123}  \\
\underline{Display} the digits 123. &  & 123 \\
Print the digits \_. &  & 1234567890 \\
Print the \underline{numbers} 123. &  & 123 \\ \midrule
\textbf{Say the word green.} & \multirow{4}{*}{Green \ding{51}} & \textbf{Green} \\
Say the word \underline{blue}. &  & Blue \\
Say the \underline{term} green. &  & Green \\
Say the word \_. &  & Hello \\ \midrule
\textbf{Repeat the term mirror.} & \multirow{4}{*}{Mirror \ding{51}} & \textbf{Mirror} \\
\underline{Recite} the term mirror. &  & Mirror \\
Repeat the term \underline{glass}. &  & Glass \\
Repeat the \underline{word} mirror. &  & Mirror \\ \midrule
\textbf{Display the word circle.} & \multirow{4}{*}{Circle \ding{51}} & Circle \\
Display the \underline{letter} circle. &  & O \\
Display the word \underline{Square}. &  & square \\
Display the word \underline{Round}. &  & round \\ \bottomrule
\end{tabular}
}
\end{minipage}%
\hfill % Ensures that the following minipage is placed right next to the previous one
\begin{minipage}[t]{0.48\textwidth} % Matching width for the second minipage
\subfloat[LIME]{
\label{subtab:table2}
\begin{tabular}{@{}lcc@{}}
\toprule
\textbf{Prompt} & \textbf{Identified MSW} & \textbf{Model Output} \\ \midrule
\textbf{Print the digits 123} & \multirow{4}{*}{{Print \ding{55}}} & \textbf{123} \\
\_ the digits 123. &  & One \\
\_ \_ digits \_. &  & Numbers \\
Print \_ \_ 123 &  & 123 \\ \midrule
\textbf{Say the word green.} & \multirow{4}{*}{Green \ding{51}} & \textbf{Green} \\
\_ \_ \_ green &  & Echo \\
Say the \_ green &  & Green \\
Say \_ \_ green &  & Green \\ \midrule
\textbf{Repeat the term mirror.} & \multirow{4}{*}{{Repeat \ding{55}}} & \textbf{Mirror} \\
\_ the term mirror. &  & Reflection \\
\_ \_ \_ mirror. &  & Reflect \\
Repeat \_ term mirror. &  & Mirror \\ \midrule
\textbf{Display the word circle.} & \multirow{4}{*}{Circle \ding{51}} & Circle \\
\_ \_ word circle. &  & Round \\
Display the word \_. &  & Do \\
Display \_ word circle. &  & Circle \\ \bottomrule
\end{tabular}
}
\end{minipage}

\caption{Comparison of GPT-4o Mini's responses to the original validation prompt (bolded) and perturbed examples (underlined alterations), using ZIP and LIME methods. Check (\ding{51}) and cross (\ding{55}) marks indicate if the Most Significant Word (MSW) was identified correctly by each method.}
\label{tab:validations_output}
\end{table*}

\subsection{ZIP Scores}
\label{sec:msw_ap}
%--------------------------------------------------------

\begin{table*}[!h]
\centering
\footnotesize
\setlength{\tabcolsep}{2pt}      % tighter column padding
\renewcommand{\arraystretch}{0.95}

% --- Left mini table ---------------------------------------------------------
\begin{minipage}[t]{0.43\textwidth}
\centering
\subfloat[Classification Tasks\label{subtab:classification_tasks2}]{
\begin{tabular}{@{}lcccccc@{}}
\toprule
& \multicolumn{2}{c}{AQUA-RAT} & \multicolumn{2}{c}{Big Bench} & \multicolumn{2}{c}{GSM8K} \\ 
\cmidrule(lr){2-3} \cmidrule(lr){4-5} \cmidrule(lr){6-7}
& Top 3 MSWs & ZIP & Top 3 MSWs & ZIP & Top 3 MSWs & ZIP \\ 
\midrule
0-CoT & \textbf{Step-by-step} & \textbf{43.94} & \textbf{Step-by-step} & \textbf{48.38} & \textbf{Step-by-step} & \textbf{18.50} \\
0-CoTB & \textbf{Problem} & \textbf{38.09} & \textbf{Step-by-step} & \textbf{38.71} & \textbf{Step-by-step} &\textbf{16.23} \\
       & Breath & 37.52 & Breath & 38.57 &  &  \\
       &  &  & Problem & 38.33 &  &  \\
0-CoTR & \textbf{Sure} & \textbf{41.04} & \textbf{Right} & \textbf{39.59} & \textbf{Step-by-step} & \textbf{16.16} \\
       &  &  & Sure & 39.52 &  &  \\
       &  &  & Let's & 38.88 &  &  \\
0-IRR & \textbf{Feel} & \textbf{56.00} & \textbf{Description} & \textbf{49.43} & \textbf{Irrelevant} & \textbf{39.74} \\
       & Description & 50.15 & Information & 48.51 & Ignore & 38.06 \\
       & Free & 50.00 & Free & 48.22 & Problem & 35.09 \\
0-PS & \textbf{Let's} & \textbf{38.72} & \textbf{Plan} & \textbf{45.17} & \textbf{Devise} & \textbf{19.73} \\
     & Carry & 38.33 & Problem & 45.16 & Plan & 19.38 \\
     & Problem & 38.33 & Let's & 45.08 &  &  \\
\bottomrule
\end{tabular}
}
\end{minipage}%
\hfill % Fills the gap between minipages
% --- Right mini table --------------------------------------------------------
\begin{minipage}[t]{0.43\textwidth}
\centering
\subfloat[Translation Tasks\label{subtab:translation_tasks2}]{
\begin{tabular}{@{}ccccc@{}}
\toprule
& \multicolumn{2}{c}{WMT 19: German} & \multicolumn{2}{c}{WMT 19: Chinese} \\ 
\cmidrule(lr){2-3} \cmidrule(lr){4-5}
& Top 3 MSWs & ZIP & Top 3 MSWs & ZIP \\ 
\midrule
0-DSP & \textbf{Step-by-step} & \textbf{11.24} & \textbf{Step-by-step} & \textbf{7.90} \\
      & Translation & 8.78 & Translation & 6.92 \\
      & Provide & 7.39 & Sentence & 6.48 \\
0-DTG & \textbf{Firstly} & \textbf{10.78} & \textbf{Firstly} & \textbf{7.34} \\
      & Detect & 10.04 & Error & 6.46 \\
      & Translation & 9.59 & Detect & 6.10 \\
\bottomrule
\end{tabular}
}
\end{minipage}
\vspace{-2mm}
\caption{Top three most significant words (MSWs) and their ZIP scores for classification and translation tasks on GPT 3.5 Turbo, with the most significant word in \textbf{bold}. All reported words are confirmed as \textit{significantly important}.}
\label{tab:Most_Important_GPT35}
\end{table*}

\clearpage

%-----------------------------------------------------

\begin{table*}[!h]
\centering
\footnotesize
\setlength{\tabcolsep}{2pt}      % tighter column padding
\renewcommand{\arraystretch}{0.95}

% --- Left mini table ---------------------------------------------------------
\begin{minipage}[t]{0.43\textwidth}
\centering
\subfloat[Classification Tasks\label{subtab:classification_tasks3}]{
\begin{tabular}{@{}lcccccc@{}}
\toprule
& \multicolumn{2}{c}{AQUA-RAT} & \multicolumn{2}{c}{Big Bench} & \multicolumn{2}{c}{GSM8K} \\ 
\cmidrule(lr){2-3} \cmidrule(lr){4-5} \cmidrule(lr){6-7}
& Top 3 MSWs & ZIP & Top 3 MSWs & ZIP & Top 3 MSWs & ZIP \\ 
\midrule
0-CoT  & \textbf{Let's} & \textbf{58.53} & \textbf{Let's} & \textbf{58.60} & - & -  \\
       & Step-by-step & 54.94 & Step-by-step & 58.50 &  &   \\
0-CoTB & \textbf{Breath} &\textbf{58.38} & \textbf{Step-by-step} & \textbf{85.09} & \textbf{Step-by-step} & \textbf{33.66} \\
       & Problem & 57.47 & Work & 57.11 & Breath & 33.23 \\
       & Take & 57.11 & Breath & 54.66 & Work & 32.66 \\
0-CoTR & \textbf{Step-by-step} & \textbf{55.22} & \textbf{Let's} & \textbf{57.24} & \textbf{Step-by-step} & \textbf{32.16}\\
       & Let's & 53.73 & Work & 57.00 & Let's & 31.95 \\
       & Way & 50.42 & Step-by-step & 55.38 & Work & 29.93 \\
0-IRR  & \textbf{Ignore} &\textbf{46.48} & \textbf{Ignore} & \textbf{24.66} & \textbf{Description} & \textbf{45.12} \\
       & Free & 45.55 & Irrelevant & 21.23 & Ignore & 43.93 \\
       & Description & 42.92 & Information & 20.76 & Irrelevant & 43.02 \\
0-PS   & \textbf{Let's} & \textbf{58.63} & \textbf{Carry} & \textbf{60.66} & \textbf{Understand} & \textbf{40.23} \\
       & Solve & 56.87 & Understand & 60.38 & First & 39.83 \\
       & First & 56.61 & First & 60.27 & Problem & 39.22 \\
\bottomrule
\end{tabular}
}
\end{minipage}%
\hfill % Fills the gap between minipages
% --- Right mini table --------------------------------------------------------
\begin{minipage}[t]{0.43\textwidth}
\centering
\subfloat[Translation Tasks\label{subtab:translation_tasks3}]{
\begin{tabular}{@{}ccccc@{}}
\toprule
& \multicolumn{2}{c}{WMT 19: German} & \multicolumn{2}{c}{WMT 19: Chinese} \\ 
\cmidrule(lr){2-3} \cmidrule(lr){4-5}
& Top 3 MSWs & ZIP & Top 3 MSWs & ZIP \\ 
\midrule
0-DSP & \textbf{Translation} & \textbf{13.39} & \textbf{Step-by-step} & \textbf{7.55} \\
      & Step-by-step & 12.25 & Translation & 2.73 \\
      & Following & 12.02 & Provide & 2.34 \\
0-DTG & \textbf{Detect} & \textbf{13.59} & \textbf{Please} & \textbf{8.00} \\
      & Please & 13.28 & Type & 7.91 \\
      & Firstly & 13.01 & Firstly & 7.89 \\
\bottomrule
\end{tabular}
}
\end{minipage}
\caption{Top three most significant words (MSWs) and their ZIP scores for classification and translation tasks on Mixtral, with the most significant word in \textbf{bold}. All reported words are confirmed as \textit{significantly important}.}
\label{tab:Most_Important_Mixtral}
\end{table*}

%-----------------------------------------------------
\begin{table*}[!h]
\centering
\footnotesize
\setlength{\tabcolsep}{2pt}      % tighter column padding
\renewcommand{\arraystretch}{0.95}

% --- Left mini table ---------------------------------------------------------
\begin{minipage}[t]{0.43\textwidth}
\centering
\subfloat[Classification Tasks\label{subtab:classification_tasks4}]{
\begin{tabular}{@{}lcccccc@{}}
\toprule
& \multicolumn{2}{c}{AQUA-RAT} & \multicolumn{2}{c}{Big Bench} & \multicolumn{2}{c}{GSM8K} \\ 
\cmidrule(lr){2-3} \cmidrule(lr){4-5} \cmidrule(lr){6-7}
& Top 3 MSWs & ZIP & Top 3 MSWs & ZIP & Top 3 MSWs & ZIP \\ 
\midrule
0-CoT  & \textbf{Let's} & \textbf{50.13} & \textbf{Think} & \textbf{35.13} & - & - \\
       & Step-by-step & 48.55 & Let's & 30.33 &  &  \\
       & Think & 46.66 & Step-by-step & 22.05 &  &  \\
0-CoTB & \textbf{Step-by-step} & \textbf{26.19} & \textbf{Take} & \textbf{81.61} & \textbf{Take} & \textbf{61.94} \\
       &  &  & Breath & 80.33 & Deep & 61.91 \\
       &  &  & Step-by-step & 80.19 & Step-by-step & 60.57 \\
0-CoTR & \textbf{Work} & \textbf{58.60} & \textbf{Let's} & \textbf{69.15} & \textbf{Let's} & \textbf{51.11} \\
       & Let's & 57.64 & Work & 69.00 & Step-by-step & 50.22 \\
       &  &  & Step-by-step & 65.72 & Work & 44.73 \\
0-IRR  & \textbf{Ignore} & \textbf{40.24} & \textbf{Feel} & \textbf{14.33} & - & - \\
       & Free & 36.88 & Ignore & 13.03 &  &  \\
       & Description & 36.35 & Free & 12.88 &  &  \\
0-PS   & \textbf{Let's} & \textbf{65.54} & \textbf{Solve} & \textbf{64.76} & \textbf{Problem} & \textbf{64.55} \\
       & Problem & 62.72 & Problem & 63.94 & Let's & 61.50 \\
       &  &  & First & 62.30 & Understand & 59.57 \\
\bottomrule
\end{tabular}
}
\end{minipage}%
\hfill % Fills the gap between minipages
% --- Right mini table --------------------------------------------------------
\begin{minipage}[t]{0.43\textwidth}
\centering
\subfloat[Translation Tasks\label{subtab:translation_tasks4}]{
\begin{tabular}{@{}ccccc@{}}
\toprule

& \multicolumn{2}{c}{WMT 19: German} & \multicolumn{2}{c}{WMT 19: Chinese} \\ 
\cmidrule(lr){2-3} \cmidrule(lr){4-5}
& Top 3 MSWs & ZIP & Top 3 MSWs & ZIP \\ 
\midrule
0-DSP & \textbf{Step-by-step} & \textbf{11.58} & \textbf{Please} & \textbf{11.15} \\
      & Translation & 9.94 & Step-by-step & 7.00 \\
      & Complete & 9.77 & Translation & 6.55 \\
0-DTG & \textbf{Refine} & \textbf{13.32} & \textbf{Please} & \textbf{5.64} \\
      & Firstly & 12.90 & Type & 5.55 \\
      & Translation & 12.65 & Firstly & 5.53 \\
\bottomrule
\end{tabular}
}
\end{minipage}
\caption{Top three most significant words (MSWs) and their ZIP scores for classification and translation tasks on Llama 2, with the most significant word in \textbf{bold}. All reported words are confirmed as \textit{significantly important}.}
\label{tab:Most_Important_Llama}
\end{table*}

\begin{table}[!htbp]
\centering
\footnotesize
\begin{tabular}{@{}lcc@{}}
\toprule
\textbf{Prompt Type} & \textbf{Correlation (r)} & \textbf{Significance (p)} \\
\midrule
0-CoT & -0.9948 & 0.064 \\
0-CoTB & -0.9954 & 0.060 \\
0-CoTR & -0.9982 & 0.037* \\
0-IRR & -0.9999 & 0.005** \\
0-PS & -0.9970 & 0.049* \\
\bottomrule
\end{tabular}
\caption{Pearson correlations between ZIP scores and GPT-4's accuracy across classification tasks for different zero-shot prompts. All correlations show strong negative relationships (r $<$ -0.99), with several reaching statistical significance (*p $<$ 0.05, **p $<$ 0.01). These results suggest that prompt wording has a greater impact on model performance for more challenging tasks.}
\label{tab:corr}
\end{table}

\clearpage

\subsection{Qualitative Analysis}
\label{sec:extended_ap}

% In your preamble, you should have:
% \usepackage{tabularx}
% \usepackage{xcolor}
% \usepackage{booktabs}

\begin{table*}[!h]
\centering
\scriptsize
\setlength{\tabcolsep}{5pt}
\renewcommand{\arraystretch}{1}
\resizebox{\textwidth}{!}{%
\begin{tabularx}{\textwidth}{%
    >{\raggedright\arraybackslash}X
    >{\raggedright\arraybackslash}X
    >{\centering\arraybackslash}p{1.5cm}
    >{\raggedright\arraybackslash}X
  }
  \toprule
  \textbf{Task Information}
  & \textbf{LLM Answer (Original Prompt)}
  & \textbf{Perturbation}
  & \textbf{LLM Answer (Perturbed Prompt)} \\

\midrule

Dorothy is 15 years old and wants to go to a museum with her family. Her family consists of her, her younger brother, her parents, and her grandfather. The regular ticket cost is \$10. People 18 years old or younger have a discount of 30\%. How much money will Dorothy have after this trip, when she currently has \$70? \newline
\textbf{Correct Answer:} 26 \newline
\textbf{Prompt:} 0-CoT \newline
\textbf{Dataset:} GSM8K
  &

1. \textbf{Identify family members}: \newline
2. \textbf{Determine ticket prices}: \newline
- Regular ticket cost: \$10 \newline
- Discounted ticket price: \$10 - \$3 = \$7 \newline
3. \textbf{Calculate the number of tickets}: \newline
   - Dorothy (15 years old): 1 ticket at \$7 \newline
   - Younger brother (10 years old): 1 ticket at \$7 \newline
   - Parents (2 adults): 2 tickets at \$10 (= \$20) \newline
   - Grandfather (1 adult): 1 ticket at \$10 \newline
4. \textbf{Calculate total cost}: ... \$44 \newline
5. \textbf{Calculate money left after the trip}: \newline
\$70 - \$44 = \$26
  &

Step-by-step

$\downarrow$

(Removed)
  &

There are 3 people eligible for the discounted ticket (Dorothy, her brother, and her grandfather) and 2 adults who will pay full price. 

1. \textbf{Discounted tickets} for Dorothy, her brother, and her grandfather: 3 $\times$ 7 = 21

2. \textbf{Regular tickets} for her parents: 2 $\times$ 10 = 20 

Now, we sum the total cost of all tickets: 21 + 20 = 41 

Next, we subtract the total ticket cost from Dorothy's current amount of money: 70 - 41 = 29
\\

\midrule

A banana costs Rs.\ 3 and an apple costs Rs.\ 4. What will be the total cost of 4 dozen apples and bananas? \newline
\textbf{Correct Answer:} 336 \newline
\textbf{Prompt:} 0-CoTB \newline
\textbf{Dataset:} AQUA-RAT
  &

1. Calculate the number of apples and bananas in 4 dozen: ... \newline
- Therefore, 4 dozen: 4 $\times$ 12 = 48 pieces of fruit. \newline
2. Determine the cost of apples and bananas: ... \newline
3. Calculate the total cost for 48 fruits: \newline
   Let's assume we buy 24 bananas and 24 apples (since we have no specific distribution given). \newline
   Cost of 24 bananas: 24 $\times$ 3 = 72 Rs. \newline
   Cost of 24 apples: 24 $\times$ 4 = 96 Rs. \newline
4. Add the costs together: \newline
   Total cost: 72 + 96 = 168 Rs.
  &

Step-by-step

$\downarrow$

(Removed)
  &

1. Cost of One Banana: Rs.\ 3 \newline
2. Cost of One Apple: Rs.\ 4 \newline
Next, we calculate the cost for 4 dozen (which is 48) of each fruit: \newline
- Cost of 4 dozen Bananas: 48 $\times$ 3 = 144 Rs.\newline
- Cost of 4 dozen Apples: 48 $\times$ 4 = 192 Rs.\newline
Now, we can add the costs of the bananas and apples together to find the total cost: \newline
144 + 192 = 336 Rs.
\\

\midrule

Mr.\ X died when his age was 1/59th of year of birth. What was his date of birth? \newline
['A)1947', 'B)1948', 'C)1980', 'D)2000', 'E)1998'] \newline
\textbf{Correct Answer:} 1947 \newline
\textbf{Prompt:} 0-PS \newline
\textbf{Dataset:} AQUA-RAT
  &

1. \textbf{Understanding the relationship}: ... This means if he was born in year $Y$ and died in year $D$, then his age at death can also be expressed as $D - Y$. \newline
2. \textbf{Setting up the equation}: $D - Y = \frac{1}{59} Y$ \newline
3. \textbf{Calculating the year of death}: Since $D$ must be a whole number, $Y$ must be a multiple of 59. \newline
4. \textbf{Finding candidates for year of birth}: The answer choices are: 1947, 1948, 1980, 2000, 1998. \newline
5. \textbf{Calculating multiples of 59}: \newline
   - The multiples of 59 around the years given are: $59 \times 33 = 1947$ \newline
   ... \newline
   Thus, the only candidate from our choices that is a multiple of 59 is 1947.
  &

\textit{Solve}

$\downarrow$

Work out
  &

1. \textbf{Understanding the relationship between age and year of birth}: ... If he was born in year $Y$ and died in year $D$, his age at death can be expressed as $D - Y$. \newline
2. \textbf{Setting up the equation}: $D - Y = \frac{1}{59} Y$ \newline
3. \textbf{Finding possible years of birth}: Since $D$ must be a whole number, $Y$ must be a multiple of 59. \newline
4. \textbf{Checking the answer choices}: \newline
   - \textbf{A: 1947}: 1947/59 = 33 (not a whole number)
   
   ... 
   
5. \textbf{Conclusion}: None of the given years are multiples of 59.
\\

\midrule

Niles is collecting his book club's annual fees. Each of the six members pays \$150/year towards snacks, plus \$30 each for six hardcover books and \$12 each for six paperback books. How much money does Niles collect in total? \newline
\textbf{True Answer:} 2412 \newline
\textbf{Prompt:} 0-IRR \newline
\textbf{Dataset:} GSM8k
  &

1. Each member pays an annual fee of \$150 for snacks. \newline
2. Each member buys 6 hardcover books at \$30 each: 6 $\times$ 30 = 180 \newline
3. Each member buys 6 paperback books at \$12 each: 6 $\times$ 12 = 72 \newline
Now, we can calculate the total amount paid by each member: 150 + 180 + 72 = 402 \newline
Next, we find the total amount collected from all 6 members: 6 $\times$ 402 = 2412 \newline
The final answer is 2412.
  &

Irrelevant

$\downarrow$

Unrelated
  &

1. Each member pays: \newline
- Annual fee for snacks: \$150 \newline
- Cost for six hardcover books: \$30 \newline
- Cost for six paperback books: \$12 \newline
Total per member: 150 + 30 + 12 = 192 \newline
2. There are 6 members, so the total amount collected from Niles is: 192 $\times$ 6 = 1152 \newline
The final answer is 1152.
\\

\bottomrule
\end{tabularx}
}
\caption{Comparison of GPT-4o mini’s partial responses to the original zero-shot prompt and a perturbed version where one of the top 3 most important words identified by ZIP was modified.}
\label{tab:zip_qualitative}
\end{table*}

\clearpage

\subsection{Human Evaluation}
\label{sec:human_eval_ap}

\begin{figure*}[!h]
    \centering
    \begin{subfigure}[b]{0.24\textwidth}
        \centering
        \includegraphics[width=\linewidth]{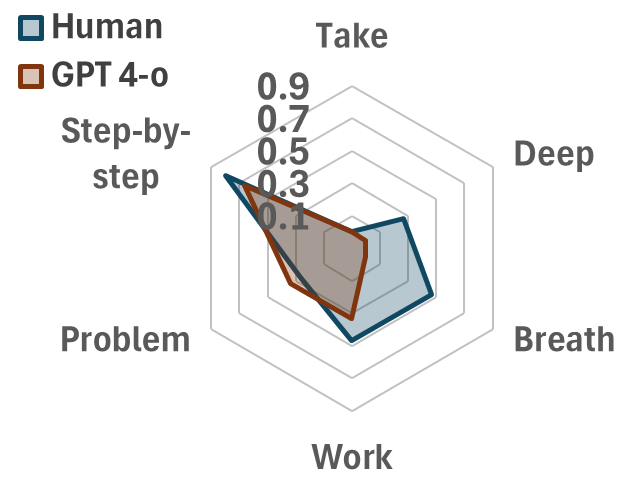}
        \caption{GPT-4o mini}
    \end{subfigure}\hfill
    \begin{subfigure}[b]{0.24\textwidth}
        \centering
        \includegraphics[width=\linewidth]{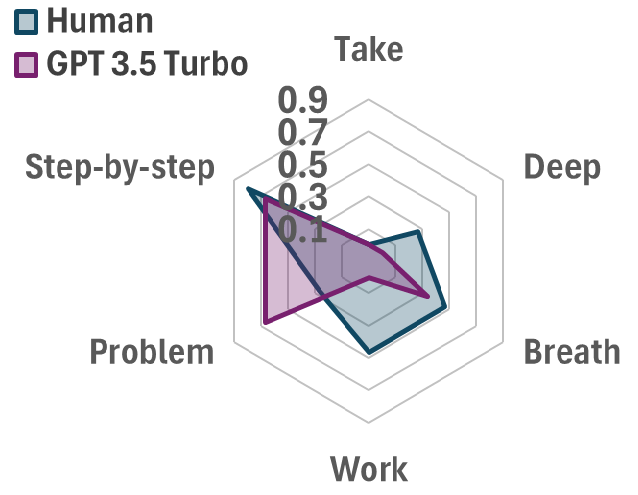}
        \caption{GPT-3.5-turbo}
    \end{subfigure}\hfill
    \begin{subfigure}[b]{0.24\textwidth}
        \centering
        \includegraphics[width=\linewidth]{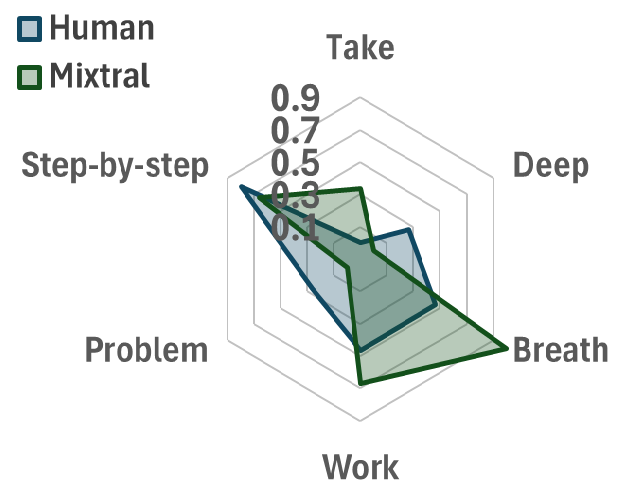}
        \caption{Mixtral}
    \end{subfigure}\hfill
    \begin{subfigure}[b]{0.24\textwidth}
        \centering
        \includegraphics[width=\linewidth]{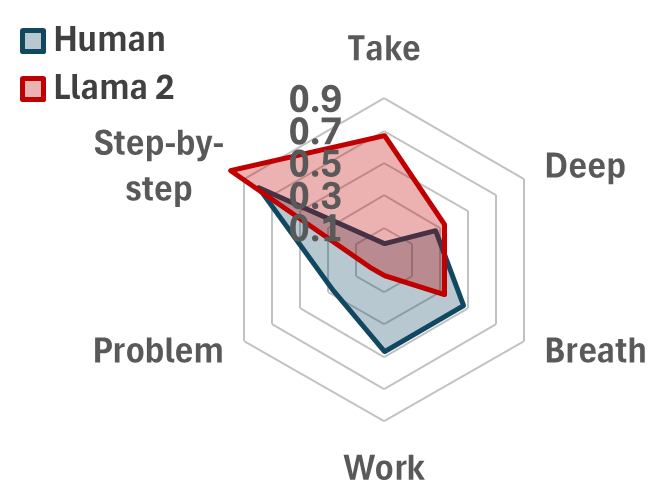}
        \caption{Llama 2}
    \end{subfigure}
    \caption{Comparison of human judgments vs model-derived word importance for the CoTB prompt.}
    \label{fig:human_eval_CoT-B}
\end{figure*}

\begin{figure*}[!h]
    \centering
    \begin{subfigure}[b]{0.24\textwidth}
        \centering
        \includegraphics[width=\linewidth]{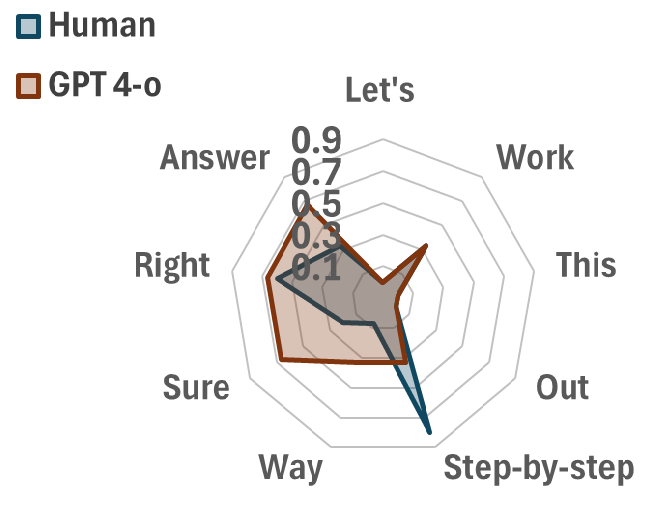}
        \caption{GPT-4o mini}
    \end{subfigure}\hfill
    \begin{subfigure}[b]{0.24\textwidth}
        \centering
        \includegraphics[width=\linewidth]{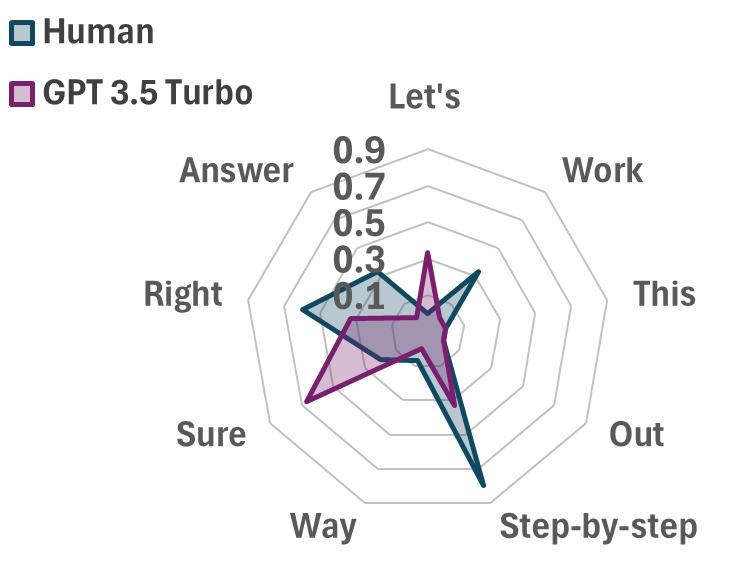}
        \caption{GPT-3.5-turbo}
    \end{subfigure}\hfill
    \begin{subfigure}[b]{0.24\textwidth}
        \centering
        \includegraphics[width=\linewidth]{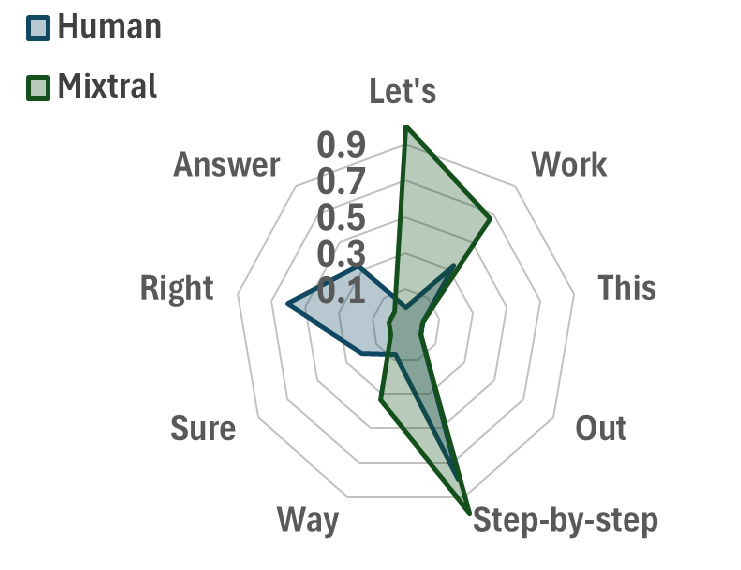}
        \caption{Mixtral}
    \end{subfigure}\hfill
    \begin{subfigure}[b]{0.24\textwidth}
        \centering
        \includegraphics[width=\linewidth]{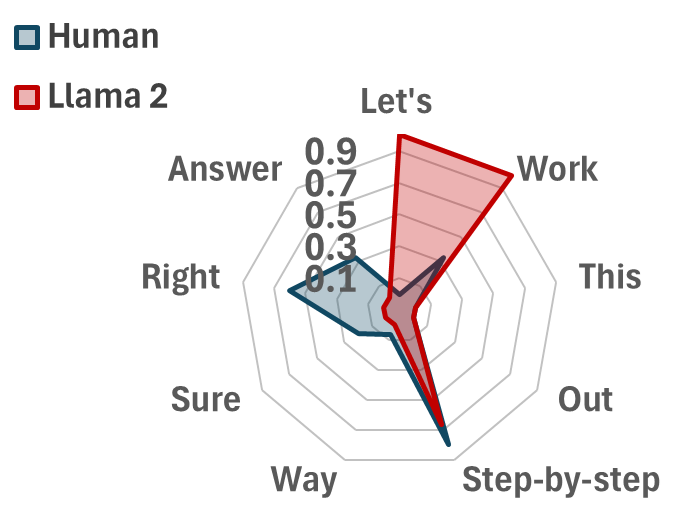}
        \caption{Llama 2}
    \end{subfigure}
    \caption{Comparison of human judgments vs model-derived word importance for the CoTR prompt.}
    \label{fig:human_eval_CoT-R}
\end{figure*}

\begin{figure*}[!h]
    \centering
    \begin{subfigure}[b]{0.24\textwidth}
        \centering
        \includegraphics[width=\linewidth]{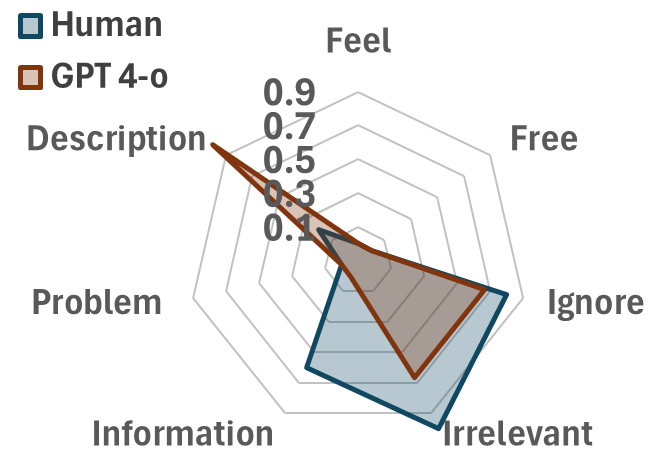}
        \caption{GPT-4o mini}
    \end{subfigure}\hfill
    \begin{subfigure}[b]{0.24\textwidth}
        \centering
        \includegraphics[width=\linewidth]{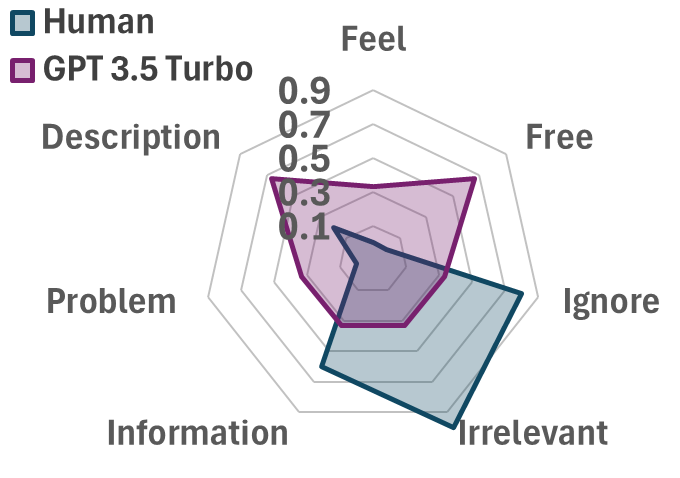}
        \caption{GPT-3.5-turbo}
    \end{subfigure}\hfill
    \begin{subfigure}[b]{0.24\textwidth}
        \centering
        \includegraphics[width=\linewidth]{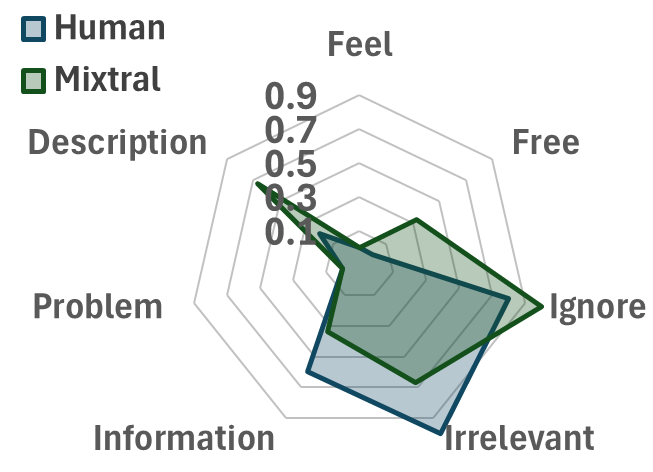}
        \caption{Mixtral}
    \end{subfigure}\hfill
    \begin{subfigure}[b]{0.24\textwidth}
        \centering
        \includegraphics[width=\linewidth]{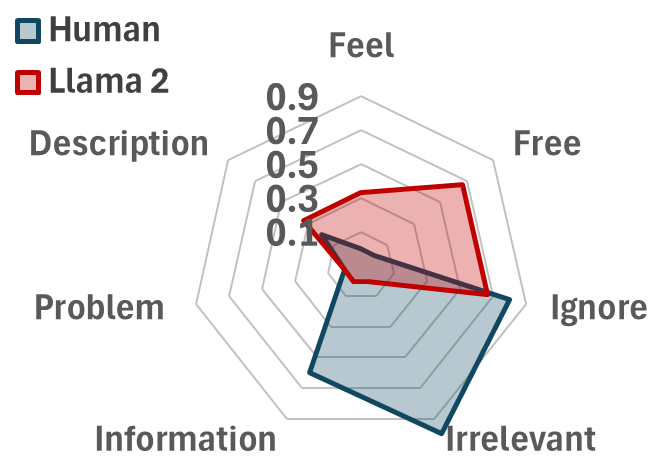}
        \caption{Llama 2}
    \end{subfigure}
    \caption{Comparison of human judgments vs model-derived word importance for the IRR prompt.}
    \label{fig:human_eval_IRR}
\end{figure*}

\begin{figure*}[!h]
    \centering
    \begin{subfigure}[b]{0.24\textwidth}
        \centering
        \includegraphics[width=\linewidth]{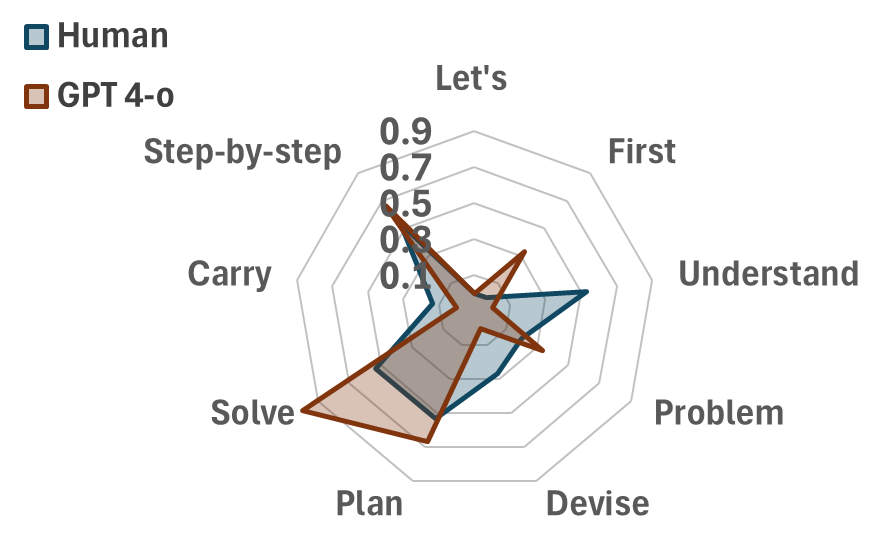}
        \caption{GPT-4o mini}
    \end{subfigure}\hfill
    \begin{subfigure}[b]{0.24\textwidth}
        \centering
        \includegraphics[width=\linewidth]{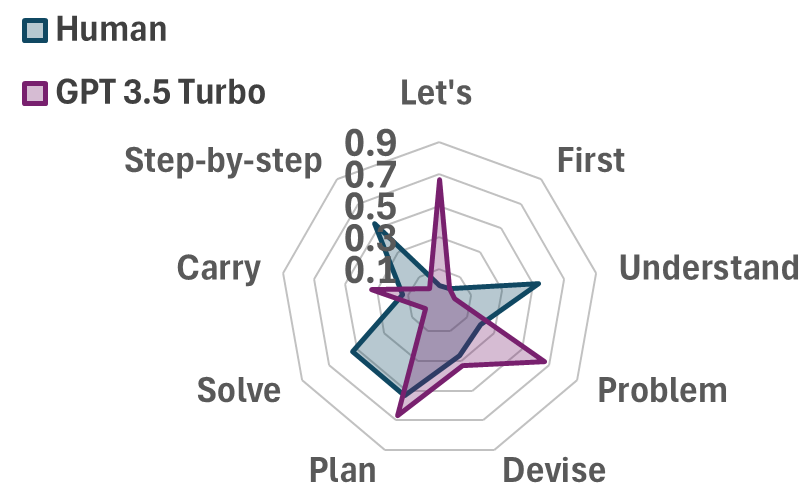}
        \caption{GPT-3.5-turbo}
    \end{subfigure}\hfill
    \begin{subfigure}[b]{0.24\textwidth}
        \centering
        \includegraphics[width=\linewidth]{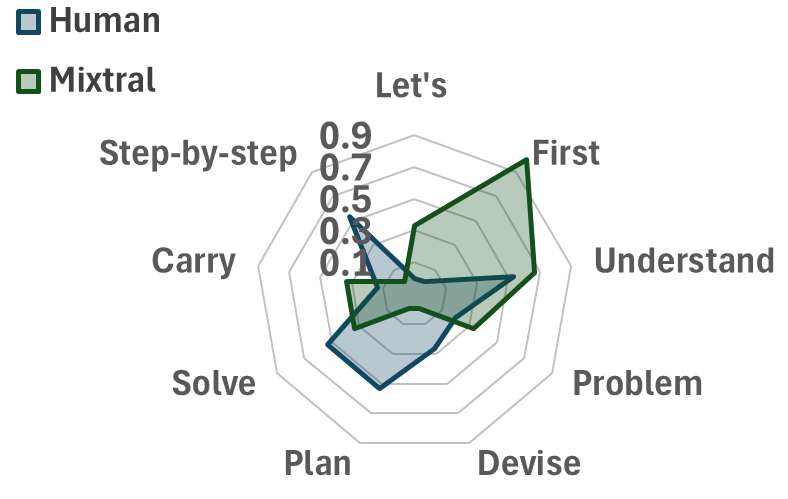}
        \caption{Mixtral}
    \end{subfigure}\hfill
    \begin{subfigure}[b]{0.24\textwidth}
        \centering
        \includegraphics[width=\linewidth]{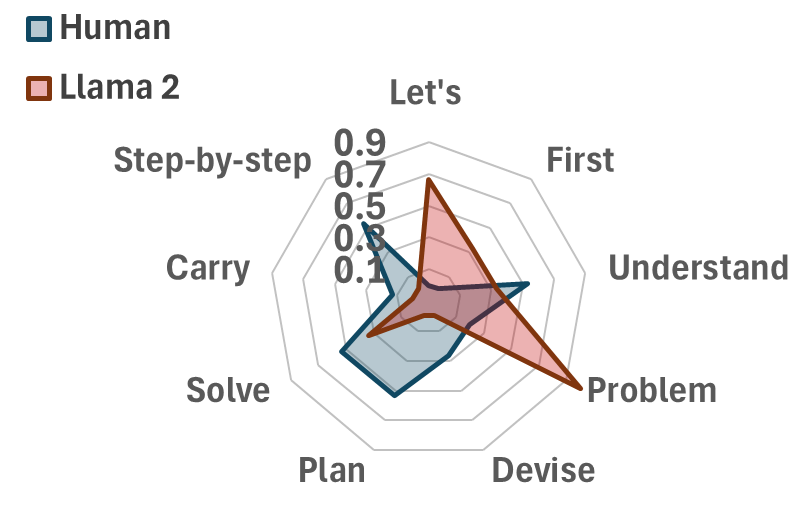}
        \caption{Llama 2}
    \end{subfigure}
    \caption{Comparison of human judgments vs model-derived word importance for the PS prompt.}
    \label{fig:human_eval_PS}
\end{figure*}

\begin{figure*}[!h]
    \centering
    \begin{subfigure}[b]{0.24\textwidth}
        \centering
        \includegraphics[width=\linewidth]{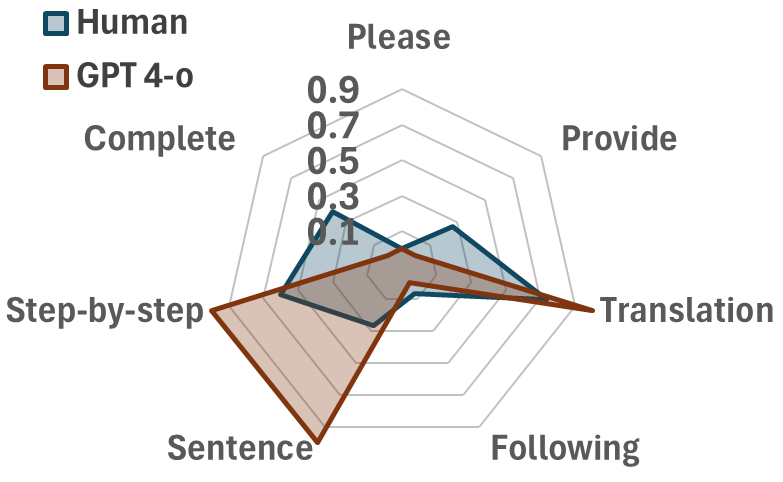}
        \caption{GPT-4o mini}
    \end{subfigure}\hfill
    \begin{subfigure}[b]{0.24\textwidth}
        \centering
        \includegraphics[width=\linewidth]{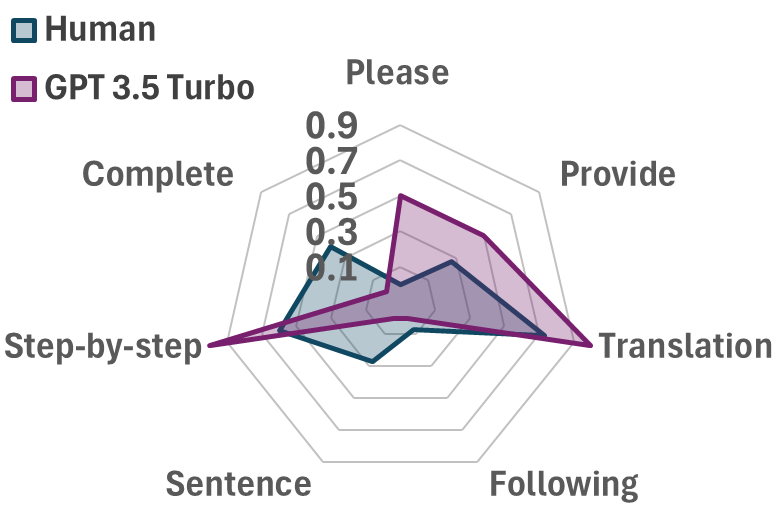}
        \caption{GPT-3.5-turbo}
    \end{subfigure}\hfill
    \begin{subfigure}[b]{0.24\textwidth}
        \centering
        \includegraphics[width=\linewidth]{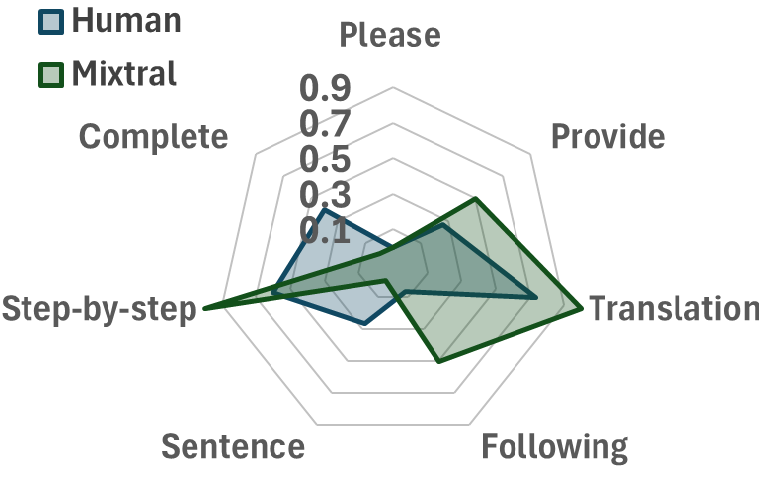}
        \caption{Mixtral}
    \end{subfigure}\hfill
    \begin{subfigure}[b]{0.24\textwidth}
        \centering
        \includegraphics[width=\linewidth]{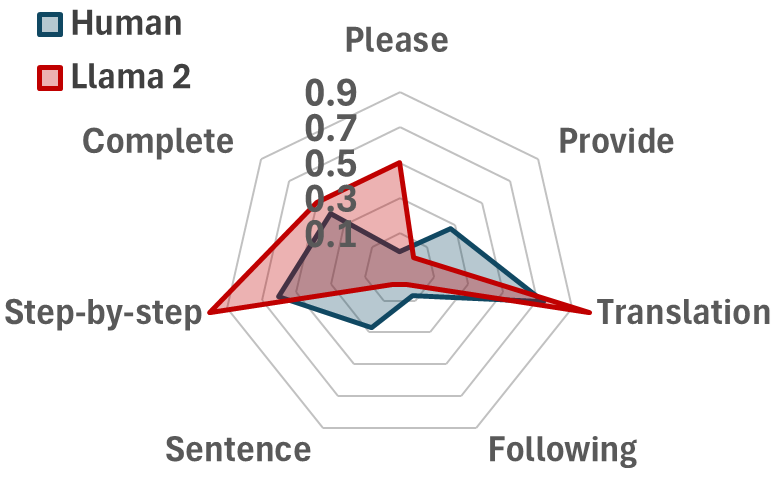}
        \caption{Llama 2}
    \end{subfigure}
    \caption{Comparison of human judgments vs model-derived word importance for the T1 prompt.}
    \label{fig:human_eval_T1}
\end{figure*}

\begin{figure*}[!h]
    \centering
    \begin{subfigure}[b]{0.24\textwidth}
        \centering
        \includegraphics[width=\linewidth]{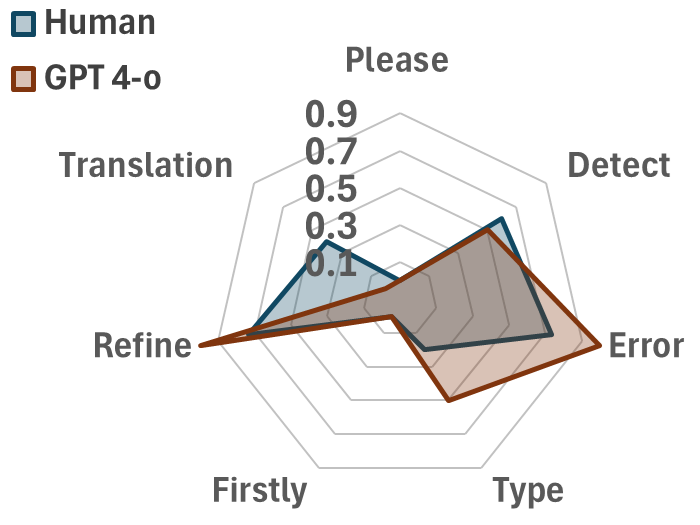}
        \caption{GPT-4o mini}
    \end{subfigure}\hfill
    \begin{subfigure}[b]{0.24\textwidth}
        \centering
        \includegraphics[width=\linewidth]{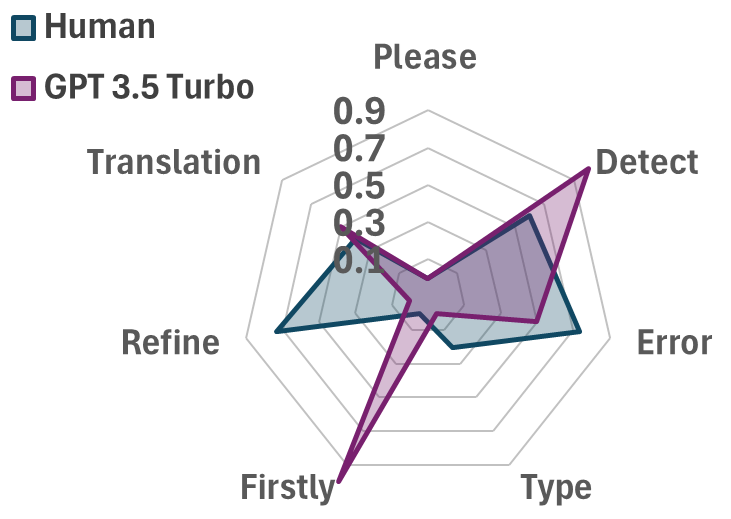}
        \caption{GPT-3.5-turbo}
    \end{subfigure}\hfill
    \begin{subfigure}[b]{0.24\textwidth}
        \centering
        \includegraphics[width=\linewidth]{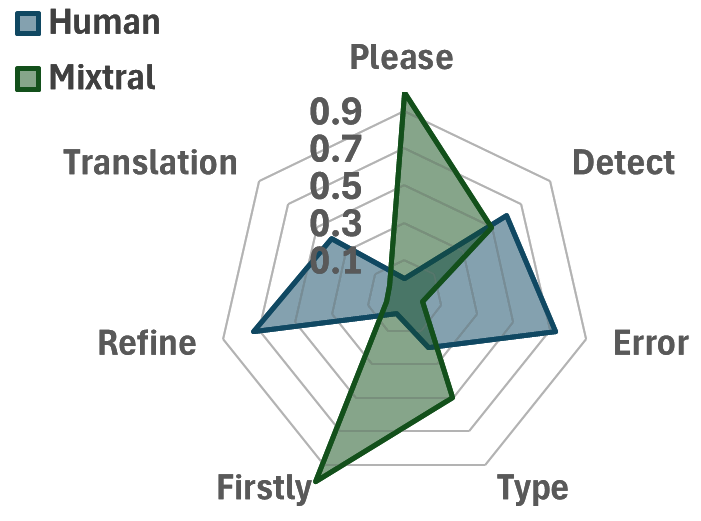}
        \caption{Mixtral}
    \end{subfigure}\hfill
    \begin{subfigure}[b]{0.24\textwidth}
        \centering
        \includegraphics[width=\linewidth]{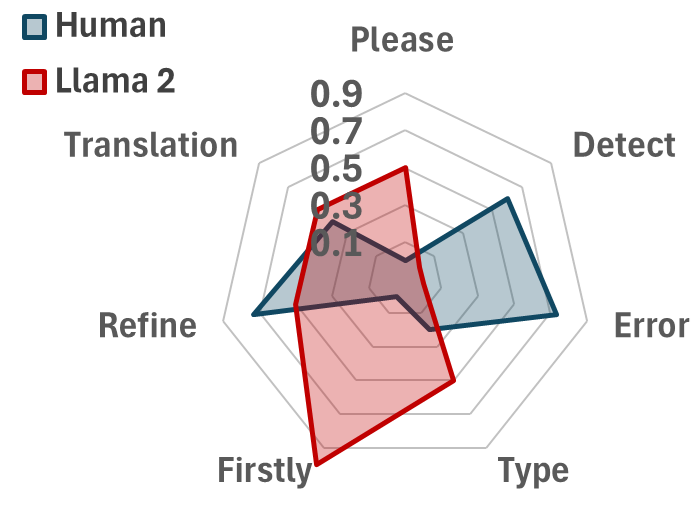}
        \caption{Llama 2}
    \end{subfigure}
    \caption{Comparison of human judgments vs model-derived word importance for the T2 prompt.}
    \label{fig:human_eval_T2}
\end{figure*}

\begin{table}[!htbp]
\centering
\renewcommand{\arraystretch}{1.1} % adds vertical space between rows

\begin{tabular}{@{}c@{\hspace{4mm}}c@{\hspace{4mm}}c@{\hspace{4mm}}c@{\hspace{4mm}}c@{\hspace{4mm}}c@{}}
\toprule
 & \textbf{0-CoT} & \textbf{0-CoTB} & \textbf{0-CoTR} & \textbf{0-IRR} & \textbf{0-PS} \\
\midrule
\textbf{Participant 1} & \makecell[c]{Think\\step-by-step} & \makecell[c]{Work\\Problem\\step-by-step} & \makecell[c]{Work} & \makecell[c]{Ignore\\Irrelevant\\Information} & \makecell[c]{Understand\\Problem\\Plan} \\ 
\textbf{Participant 2} & \makecell[c]{step-by-step} & \makecell[c]{step-by-step} & \makecell[c]{step-by-step\\Right\\Answer} & \makecell[c]{Ignore\\Irrelevant} & \makecell[c]{Devise\\Plan\\step-by-step} \\

\textbf{Participant 3} & \makecell[c]{Think} & \makecell[c]{Work} & \makecell[c]{Right} & \makecell[c]{Irrelevant} & \makecell[c]{Solve} \\

\textbf{Participant 4} & \makecell[c]{step-by-step} & \makecell[c]{step-by-step} & \makecell[c]{step-by-step} & \makecell[c]{Irrelevant} & \makecell[c]{step-by-step} \\

\textbf{Participant 5} & \makecell[c]{step-by-step} & \makecell[c]{Breath\\step-by-step} & \makecell[c]{step-by-step\\Sure\\Right} & \makecell[c]{Ignore\\Irrelevant\\Description} & \makecell[c]{Understand\\Plan\\Solve} \\
\bottomrule
\end{tabular}
\caption{Important words identified by participants across five zero-shot prompts for classification tasks.}
\label{tab:human_QA}
\end{table}

\clearpage

\begin{figure}[H]
    \centering
    \setlength{\fboxsep}{0pt} % Adjust padding inside the box, set to 0 if no padding is needed
    \setlength{\fboxrule}{1pt} % Thickness of the border
    \fbox{\includegraphics[width=\textwidth]{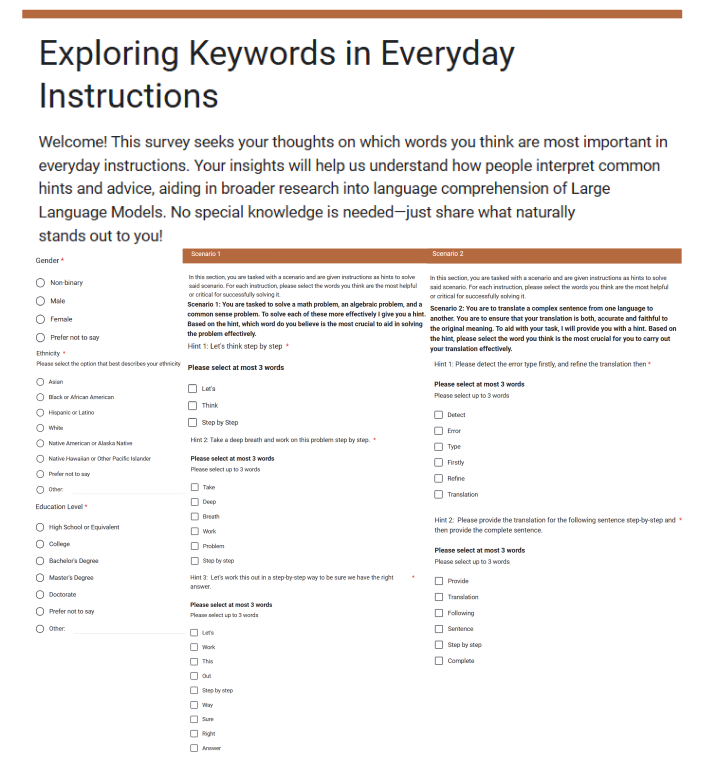}}
    \caption{Google evaluation form for studying human intuition on word importance.}
    \label{fig:human_eval}
\end{figure}

%\vspace{1mm}

\clearpage 
%\FloatBarrier
\subsection{Prompt Templates}
\label{sec:template_ap}

\begin{table*}[!h]
\centering
\footnotesize  % Keep small font for readability
\setlength{\tabcolsep}{10pt}  % Adjust space between columns
\renewcommand{\arraystretch}{1.4}  % Set row spacing for better readability
\resizebox{\textwidth}{!}{%
\begin{tabular}{p{12cm}}  % Use a single-column layout with the same format as the previous table
\toprule
\rowcolor[HTML]{EFEFEF} \textbf{Synonym Generation Prompts} \\
\midrule
  {[}Ex: 1, Ex: 2, ..., Ex: n{]} ← Few-shot examples applied for context.\\ \\
  \textbf{Original Sentence}: {[}Instructional Prompt{]}\\ 
  \textbf{Target word}: {[}Target Word{]}\\ \\
  \textbf{Task}: Please provide 10 different meaningful alterations of the original sentence.\\ Each time replacing the word {[}Target Word{]} with a different synonym. Ensure the rest of the sentence remains unchanged.\\ Write down the altered sentence and the replaced word as the output.\\ \\
  \textbf{Output}: \\ 
\midrule
\rowcolor[HTML]{EFEFEF} \textbf{Co-hyponym Generation Prompts} \\
\midrule
  {[}Ex: 1, Ex: 2, ..., Ex: n{]} ← Few-shot examples applied for context.\\ \\
  \textbf{Original Sentence}: {[}Instructional Prompt{]}\\ 
  \textbf{Target word}: {[}Target Word{]}\\ \\
  \textbf{Task}: Please provide 10 different meaningful co-hyponyms of the original sentence.\\ Each time, replacing the word {[}Target Word{]} with a different co-hyponym. Ensure the rest of the sentence remains unchanged.\\ Write down the altered sentence and the replaced word as the output.\\ \\
  \textbf{Output}: \\ 
\midrule
\rowcolor[HTML]{EFEFEF} \textbf{Meaningfulness and Correctness Prompts} \\
\midrule
  Is this sentence meaningful and grammatically correct? “[Perturbed prompt]”\\ \\
  Answer only with Yes or No.\\
\bottomrule
\end{tabular}%
}
\caption{Prompts used for creating perturbations. This task was preceded by a few-shot example set to guide the model in generating contextually relevant synonyms.}
\label{tab:prompt_types_pert}
\end{table*}

\clearpage

\begin{table*}[!h]
\centering
\small
\setlength{\tabcolsep}{10pt}
\renewcommand{\arraystretch}{1.3}
\resizebox{\textwidth}{!}{%
\begin{tabular}{>{\textsf}p{5cm} p{10cm}}  % Adjusted column widths
\toprule
\rowcolor[HTML]{EFEFEF} \multicolumn{2}{l}{\textbf{Classification-Based Tasks}} \\
\midrule
\textbf{Task} & \textbf{Prompt template} \\
\midrule
GSM8k (0-CoT, 0-CoTB, 0-CoTR, 0-IRR, 0-PS) & {[}Question{]}. {[}Instructional Prompt{]} \\
      & Write down your final answer to the question in this format: ``The final answer is X.'' \\
      & The type of X should be a number. \\
\midrule
AQUA (0-CoT, 0-CoTB, 0-CoTR, 0-IRR, 0-PS) & Multiple-Choice Question: {[}Question{]} \\
      & Answer Choices: {[}Options{]} \\
      & {[}Instructional Prompt{]} \\
      & Write down your final answer in the format: ``The correct answer is {[}X{]}.'' where X is the letter of the correct answer choice (A, B, C, D, or E). \\
\midrule
Big Bench (0-CoT, 0-CoTB, 0-CoTR, 0-IRR, 0-PS) & Multiple-Choice Question: {[}Question{]} \\
          & Answer Choices: {[}Options{]} \\
          & {[}Instructional Prompt{]} \\
          & Write down your final answer in the format: ``The correct answer is (X).'' where X is the letter of the correct answer choice (A, B, C, D, or E). \\
\toprule
\rowcolor[HTML]{EFEFEF} \multicolumn{2}{l}{\textbf{Translation-Based Tasks}} \\
\midrule
\textbf{Task} & \textbf{Prompt template} \\
\midrule
Translation (0-DSP) & {[}Instructional Prompt{]}. {[}Original Sentence{]} \\
                & Before you write down the final English translation, please use these exact words: ``\#\#\#\#The final English translation of the complete sentence is:'' \\
\midrule
Translation (0-DTG) & Step 1: Given the sentence: {[}Original Sentence{]}, what is the English translation? \\
                & Before you write down the final English translation, please use these exact words: ``\#\#\#\#The final English translation of the complete sentence is:'' \\
                & Step 2: Given the sentence: {[}Original Sentence{]}, the English translation is {[}LLM Translation from Step 1{]}. \\
                & {[}Instructional Prompt{]}. Before you write down the final English translation, please use these exact words: ``\#\#\#\#The final English translation of the complete sentence is:'' \\
\bottomrule
\end{tabular}%
}
\caption{Prompt templates used for both classification and translation tasks across various datasets.}
\label{tab:task_prompt}
\end{table*}

%\begin{figure*}[h]
%  \centering
%  \includegraphics[width=0.9\textwidth]{Images/most_gpt35_translation.png}
%  \caption{Saliency results for translation task in German and Chinese on GPT 3.5 Turbo. This table summarizes the most significant words and the top three most significant words for seven instructional prompts. All of the reported words are confirmed as significantly significant.}
%  \label{fig:Translation_Most_significant_GPT35}
%\end{figure*}

%\FloatBarrier
\clearpage
\subsection{Perturbations}
\label{sec:pert_ap}

\begin{table}[h!]

\resizebox{\textwidth}{!}{%
\begin{tabular}{lllll}
\hline
  \vspace{2mm}
  
\textbf{\begin{tabular}[c]{@{}l@{}}Zero-shot \\  
Instructional Prompt\end{tabular}} &
  \textbf{Generated Candidates} &
  \textbf{Semantic Similarity \textgreater{}30\%} &
  \textbf{\begin{tabular}[c]{@{}l@{}}Meaningful and \\ grammatically correct\end{tabular}} &
  \textbf{Final Perturbations} \\ \hline

    \vspace{2mm}
    
Let’s think \textbf{step-by-step.} &
  \begin{tabular}[c]{@{}l@{}}Let’s think \textbf{slowly}.\\  \\  Let’s think. (removal)\\    \\ Let’s think \textbf{bit-by-bit}.\\    \\ Let’s think \textbf{piecemeal}.\end{tabular} &
  \begin{tabular}[c]{@{}l@{}}Let’s think \textbf{\underline{slowly}}.\\    \\ Let’s \textbf{think}.\\    \\ Let’s think \textbf{bit-by-bit}.\\    \\ Let’s think \textbf{piecemeal}.\end{tabular} &
  \begin{tabular}[c]{@{}l@{}}\\  \\ Let’s \textbf{think}.\\    \\ Let’s think \textbf{bit-by-bit}.\\    \\ \underline{Let’s think \textbf{piecemeal}}.\end{tabular} &
  \begin{tabular}[c]{@{}l@{}}Let’s think.\\    \\ Let’s think \textbf{bit-by-bit}.\end{tabular} \\ \hline

    \vspace{2mm}
    
\begin{tabular}[c]{@{}l@{}}Take a deep breath and \\ \textbf{work} on this problem \\ step-by-step.\end{tabular} &
  \begin{tabular}[c]{@{}l@{}}Take a deep breath and \textbf{focus}\\  on this problem step-by-step.\\    \\ Take a deep breath and \\ on this problem step-by-step. \\ (removal)\\    \\ Take a deep breath and \textbf{reflect} \\ on this problem step-by-step.\\    \\ Take a deep breath and \textbf{study} \\ on this problem step-by-step.\end{tabular} &
  \begin{tabular}[c]{@{}l@{}}Take a deep breath and \textbf{focus} \\ on this problem step-by-step.\\    \\ Take a deep breath and \\ on this problem step-by-step.\\    \\ Take a deep breath and \underline{\textbf{reflect}} \\ on this problem step-by-step.\\    \\ Take a deep breath and \textbf{study}\\ on this problem step-by-step.\end{tabular} &
  \begin{tabular}[c]{@{}l@{}}Take a deep breath and \textbf{focus} \\ on this problem step-by-step.\\    \\ \underline{Take a deep breath and} \\ \underline{on this problem step-by-step.}\\    \\ \\ \\ \\ Take a deep breath and \textbf{study} \\ on this problem step-by-step.\end{tabular} &
  \begin{tabular}[c]{@{}l@{}}Take a deep breath and \textbf{focus}\\ on this problem step-by-step.\\    \\  \\    \\ \\ \\ \\ \\ Take a deep breath and \textbf{study} \\ on this problem step-by-step.\end{tabular} \\ \hline

  \vspace{2mm}
  
%\begin{tabular}[c]{@{}l@{}}Let’s work this out in a \\ step-by-step way\\ to be \textbf{sure} we have the\\ right answer.\end{tabular} &
%\begin{tabular}[c]{@{}l@{}}Let’s work this out in a \\ step-by-step way\\ to be \textbf{secured} we have the \\ right answer.\\    \\ Let’s work this out in a \\ step-by-step way\\ to be we have the \\ right answer. (removal)\\    \\ Let’s work this out in a \\ step-by-step way \\ to be \textbf{concrete} we have the \\ right answer.\\    \\ Let’s work this out in a \\ step-by-step way \\ to be \textbf{certain} we have the \\ right answer.\end{tabular} &
%  \begin{tabular}[c]{@{}l@{}}Let’s work this out in a \\ step-by-step way \\ to be \textcolor{red}{\textbf{secured}} we have the \\ right answer.\\    \\ Let’s work this out in a \\ step-by-step way \\ to be we have the \\ right answer. \\    \\ Let’s work this out in a \\ step-by-step way \\ to be \textcolor{red}{\textbf{concrete}} we have the \\ right answer.\\    \\ Let’s work this out in a \\ step-by-step way \\ to be \textbf{certain} we have the \\ right answer.\end{tabular} &
%  \begin{tabular}[c]{@{}l@{}}\\ \\ \\ \\ \\ Let’s work this out in a \\ step-by-step way\\ to be we have the \\ right answer. \\    \\  \\    \\ \\ \\ \\ Let’s work this out in a \\ step-by-step way\\ to be certain we have the \\ right answer.\end{tabular} &
%  \begin{tabular}[c]{@{}l@{}}\\ \\ \\ \\ \\ \\ \\ \\ \\ \\ \\ \\ \\ \\ \\ Let’s work this out in a \\ step-by-step way \\ to be certain we have the \\ right answer.\end{tabular} \\ \hline
\begin{tabular}[c]{@{}l@{}}Feel free to ignore \textbf{irrelevant} \\ information in the problem \\ description.\end{tabular} &
  \begin{tabular}[c]{@{}l@{}}Feel free to ignore \textbf{insignificant} \\ information in the problem \\ description.\\    \\ Feel free to ignore\\ information in the problem \\ description. (removal)\\    \\ Feel free to ignore \textbf{irrelative} \\ information in the problem \\ description.\\    \\ Feel free to ignore \textbf{unimportant} \\ information in the problem \\ description.\end{tabular} &
  \begin{tabular}[c]{@{}l@{}}Feel free to ignore \textbf{insignificant}\\ information in the problem \\ description.\\    \\ Feel free to ignore\\ information in the problem \\ description. \\    \\ Feel free to ignore \underline{\textbf{irrelative}} \\ information in the problem \\ description.\\    \\ Feel free to ignore \textbf{unimportant} \\ information in the problem \\ description.\end{tabular} &
  \begin{tabular}[c]{@{}l@{}}Feel free to ignore \textbf{insignificant} \\ information in the problem \\ description.\\    \\ Feel free to ignore information \\ in the problem description. \\ \\ \\ \\ \\ \\ \\ Feel free to ignore \textbf{unimportant}\\ information in the problem \\ description.\end{tabular} &
  \begin{tabular}[c]{@{}l@{}}Feel free to ignore \textbf{insignificant} \\ information in the problem \\ description.\\    \\ Feel free to ignore information\\ in the problem description. \\ \\ \\ \\ \\ \\ \\ Feel free to ignore \textbf{unimportant}\\ information in the problem \\ description.\end{tabular} \\ \hline

    \vspace{2mm}

\begin{tabular}[c]{@{}l@{}}Please \textbf{detect} the error type \\ firstly and refine the \\ translation then.\end{tabular} &
  \begin{tabular}[c]{@{}l@{}}Please \textbf{observe} the error type \\ firstly and refine the \\ translation then.\\    \\ Please the error type \\ firstly and refine the \\ translation then. (removal)\\    \\ Please \textbf{notice} the error type \\ firstly and refine the \\ translation then.\\    \\ Please \textbf{discern} the error type \\ firstly and refine \\ the translation then.\\    \\ Please \textbf{pick out} the error type \\ firstly and refine \\ the translation then.\end{tabular} &
  \begin{tabular}[c]{@{}l@{}}Please \textbf{observe} the error type \\ firstly and refine\\ the translation then.\\    \\ Please the error type \\ firstly and refine the \\ translation then. \\    \\ Please \textbf{notice} the error type \\ firstly and refine the\\ translation then.\\   \\ Please \textbf{discern} the error type\\ firstly and refine the \\ translation then.\\    \\ Please \underline{\textbf{pick out}} the error type \\ firstly and refine the \\ translation then.\end{tabular} &
  \begin{tabular}[c]{@{}l@{}}Please \textbf{observe} the error type \\ firstly and refine \\ the translation then.\\    \\ \underline{Please the error type} \\ \underline{firstly and refine the}\\ \underline{translation then}. \\    \\ Please \textbf{notice} the error type \\ firstly and refine the \\ translation then.\\    \\ Please \textbf{discern} the error type \\ firstly and refine the \\ translation then.\\ \\ \\ \\ \\ \end{tabular} &
  \begin{tabular}[c]{@{}l@{}}Please \textbf{observe} the error type \\ firstly and refine \\ the translation then.\\    \\  \\    \\ \\ \\ Please \textbf{notice} the error type \\ firstly and refine the \\ translation then.\\    \\ Please \textbf{discern} the error type \\ firstly and refine the \\ translation then.\\ \\ \\ \\  \\ \end{tabular} \\ \hline

    \vspace{2mm}
  
\end{tabular}%
}
\caption{
Illustration of the multi-stage filtering of prompt perturbations for classification prompts. Each candidate is generated via synonym, co-hyponym, or removal, and filtered for semantic similarity ($>$30\%) and grammaticality. \underline{Underlined} candidates indicate perturbations that were rejected at a given filtering stage. The final column shows the valid perturbations used for evaluation.
}

\label{tab:pert_ex_classification}
\end{table}

\FloatBarrier

\begin{table}[!h]
\centering
\footnotesize
\begin{tabular}{@{}llc cccc@{}}
\toprule
Original Word & Generated Candidates & 20\% & 30\% & 40\% & 50\% \\ \midrule
\multirow{3}{*}{Let's} & We should (49.13\%) & Accept & Accept & Accept & \textbf{\underline{Reject}} \\
                       & It is recommended that we (21.29\%) & \textbf{\underline{Accept}} & Reject & Reject & Reject \\
                       & We can (51.57\%) & Accept & Accept & Accept & Accept \\
\midrule
\multirow{2}{*}{First} & Before anything else (47.05\%) & Accept & Accept & Accept & \textbf{\underline{Reject}} \\
                       & Right off the bat (25.66\%) & \textbf{\underline{Accept}} & Reject & Reject & Reject \\
\midrule
\multirow{2}{*}{Understand} & Perceive (58.85\%) & Accept & Accept & Accept & Accept \\
                            & Apprehend (42.68\%) & Accept & Accept & Accept & \textbf{\underline{Reject}} \\
\midrule
\multirow{3}{*}{Problem} & Dilemma (72.35\%) & Accept & Accept & Accept & Accept \\
                         & Hurdle (35.48\%) & Accept & Accept & \textbf{\underline{Reject}} & \textbf{\underline{Reject}} \\
                         & Difficulty (60.58\%) & Accept & Accept & Accept & Accept \\
\midrule
\multirow{3}{*}{Devise} & Design (37.65\%) & Accept & Accept & \textbf{\underline{Reject}} & \textbf{\underline{Reject}} \\
                        & Draft (36.57\%) & Accept & Accept & \textbf{\underline{Reject}} & \textbf{\underline{Reject}} \\
                        & Set up (36.85\%) & Accept & Accept & \textbf{\underline{Reject}} & \textbf{\underline{Reject}} \\
\midrule
\multirow{2}{*}{Plan}   & Procedure (44.23\%) & Accept & Accept & Accept & \textbf{\underline{Reject}} \\
                        & Strategy (57.06\%) & Accept & Accept & Accept & Accept \\
\midrule
\multirow{2}{*}{Solve}  & Crack (29.82\%) & Accept & \textbf{\underline{Reject}} & \textbf{\underline{Reject}} & \textbf{\underline{Reject}} \\
                        & Tackle (40.75\%) & Accept & Accept & Accept & \textbf{\underline{Reject}} \\
\midrule
\multirow{3}{*}{Step-by-step} & Progressively (38.88\%) & Accept & Accept & \textbf{\underline{Reject}} & \textbf{\underline{Reject}} \\
                              & Phase by phase (30.38\%) & Accept & Accept & \textbf{\underline{Reject}} & \textbf{\underline{Reject}} \\
                              & Inch by inch (21.12\%) & \textbf{\underline{Accept}} & Reject & Reject & Reject \\
\midrule
\multicolumn{2}{c}{\textbf{Accuracy}} & 85\% & \textbf{95\%} & 65\% & 40\% \\
\bottomrule
\end{tabular}
\caption{Evaluation of semantic similarity thresholds (20\%–50\%) on the 0-PS prompt using 20 manually validated word variants. Each row lists an original word, its generated candidate replacements (with similarity scores), and whether the candidate is accepted or rejected at each threshold. \textbf{\underline{Bold and underlined}} entries indicate incorrect acceptance/rejection (contradict human judgment). The accuracy row shows the overall agreement rate with human judgments for each threshold. The 30\% threshold yielded the best performance (95\% accuracy), which is why it was chosen for the main experiments.}
\label{emperical_test}
\end{table}

\FloatBarrier

\subsection{Token Usage per Instructional Prompt}
\label{sec:computation_ap}

\begin{table}[!h]
\centering
\resizebox{\textwidth}{!}{% Resize table to fit within page width
\renewcommand{\arraystretch}{1.5} % Adds padding in the cells for better readability
\begin{tabular}{@{}lcccccccccc@{}}
\toprule
& \multicolumn{3}{c}{AQUA-RAT} & \multicolumn{3}{c}{Big Bench} & \multicolumn{3}{c}{GSM8K} \\ 
\cmidrule(lr){2-4} \cmidrule(lr){5-7} \cmidrule(lr){8-10}
& Input tokens & Output tokens & Total tokens & Input tokens & Output tokens & Total tokens & Input tokens & Output tokens & Total tokens \\ 
\midrule
0-CoT  & 600765 & 1825656 & 2426421 & 871710  & 1204340 & 2076050 & 488985 & 1098887 & 1587872 \\
0-CoTB & 1735827 & 3800599 & 5536426 & 2464368  & 2502721 & 4967089 & 1435263 & 3207869 & 4643132 \\
0-CoTR & 2472444 & 4488703 & 6961147 & 3447846  & 2974113 & 6421959 & 2070036 & 4367080 & 6437116 \\
0-IRR  & 1390389 & 3014827 & 4405216 & 1974426  & 1831424 & 3805850 & 1149441 & 1895271 & 3044712 \\
0-PS   & 4088076 & 9379794 & 13467870 & 5581284  & 6834640 & 12415924 & 3472044 & 7837170 & 11309214 \\
\midrule
Total  & 10287501 & 22509579 & 32797080 & 14339634 & 15347238 & 29686872 & 8615769 & 18406277 & 27022046 \\
\bottomrule
\end{tabular}
}
\caption{Estimated token usage per instructional prompt across datasets using GPT-4o mini. Input, output, and total tokens are approximated using space-based tokenization and may differ from GPT-4o mini's actual method.}
\label{tab:computations}
\end{table}

\end{document}